\documentclass{article}
\usepackage{arxiv}
\usepackage{graphicx} 
\usepackage{hyperref}
\hypersetup{
    colorlinks=true,
    linkcolor=cyan,
    filecolor=magenta,      
    urlcolor=purple,
    citecolor = blue,
    pdftitle={Overleaf Example},
    }
\usepackage[dvipsnames,table]{xcolor}
\usepackage{multirow}
\usepackage{tabularx}
\usepackage{booktabs}
\usepackage[T1]{fontenc}
\usepackage[utf8]{inputenc}
\DeclareUnicodeCharacter{0301}{\'{}} 

\usepackage[english]{babel}
\addto\extrasenglish{

}
\usepackage{csquotes}

\usepackage{cleveref}

\usepackage[
  backend=biber,
  style=authoryear-comp,
  sortcites=true,
  natbib=true,
  giveninits=true,
  maxcitenames=2,
  maxbibnames=99,
  doi=false,
  url=true,
  isbn=false,
  dashed=false,
  sorting=ynt,
  backref=true,
]{biblatex}

\definecolor{mygray}{gray}{0.9}
\definecolor{myblue}{RGB}{100,143,255}
\definecolor{mypink}{RGB}{220,38,127}
\definecolor{mygreen}{RGB}{254,97,0}

\title{Best Practices For Empirical Meta-Algorithmic Research: Guidelines from the COSEAL Research Network}
\author{Theresa Eimer$^{1,2}$ \and Lennart Schäpermeier$^3$ \and André Biedenkapp$^4$ \and Alexander Tornede \and Lars Kotthoff$^5$ \and Pieter Leyman$^6$ \and Matthias Feurer$^{7,8}$\and Katharina Eggensperger$^{7,8}$ \and Kaitlin Maile$^9$ \and Tanja Tornede \and Anna Kozak$^{10}$ \and Ke Xue$^{11}$ \and Marcel Wever$^{1,2}$ \and Mitra Baratchi$^{12}$ \and Damir Pulatov$^{13}$ \and Heike Trautmann$^{14}$ \and Haniye Kashgarani$^{15}$ \and Marius Lindauer$^{1,2}$}
\date{\small $^1$Leibniz University Hannover, $^2$L3S Research Center, $^3$University of Münster, $^4$University Freiburg, $^5$University of St. Andrews, $^6$Ghent University, $^7$TU Dortmund University, $^8$Lamarr Institute for Machine Learning and Artificial Intelligence, $^9$Google, $^{10}$Warsaw University of Technology, $^{11}$Nanjing University, $^{12}$Leiden University, $^{13}$University of North Carolina Wilmington, $^{14}$Paderborn University, $^{15}$Purdue University}

\begin{document}
\maketitle

\begin{abstract}
    Empirical research on meta-algorithmics, such as algorithm selection, configuration, and scheduling, often relies on extensive and thus computationally expensive experiments.
    With the large degree of freedom we have over our experimental setup and design comes a plethora of possible error sources that threaten the scalability and validity of our scientific insights.
    Best practices for meta-algorithmic research exist, but they are scattered between different publications and fields, and continue to evolve separately from each other.
    In this report, we collect good practices for empirical meta-algorithmic research across the subfields of the COSEAL community, encompassing the entire experimental cycle: from formulating research questions and selecting an experimental design, to executing experiments, and ultimately, analyzing and presenting results impartially.
    It establishes the current state-of-the-art practices within meta-algorithmic research and serves as a guideline to both new researchers and practitioners in meta-algorithmic fields.
\end{abstract}

\section{Improving Experimental Design in Meta-Algorithmic Research}

Empirical meta-algorithmic research allows for a large degree of control over the experimental setup due to the computational nature of our experiments. 
However, this is a double-edged sword for research that is both meaningful and efficient: We encounter a plethora of decisions that \emph{must be made deliberately} every step of the way.
This can lead to divergent research practices and results, which in turn can diminish the amount of meaningful progress made in a field, as common standards for what constitutes reliable research results are important for long-term progress - compare \citet{aranha-swarm22,rajwar-air23} for the result of too much variation in a research field with too little systematic comparison.

In addition, we increasingly need to consider the impact our research has on other disciplines and society at large.
Practices to ensure fairness and accountability in application~\citep{EUExpert19,Kaur22,TAILORRoadmap22}, are not yet well integrated into meta-algorithmic research.
Therefore, conducting meta-algorithmic research with reliable and reproducible outcomes for the different dimensions relevant for real-world applications is an ever-increasing challenge.


\paragraph{Our Goals} are to provide a written account of ``common knowledge'' principles and best practices from the COSEAL\footnote{COSEAL (\textbf{CO}nfiguration and \textbf{SE}lection of \textbf{AL}gorithms) is an international research community on meta-algorithmic topics with more than a decade of experience on empirical research; see also \url{https://www.coseal.net/}.}  community to improve the experimental design in meta-algorithmic research in order to promote more research that matters long-term for real-life applications.
Specifically, we focus on:
\begin{enumerate}
    \item Drawing empirically sound conclusions through carefully designed evaluation pipelines.
    \item Improving efficiency of research workflows to enable more and better scientific insights throughout the field.
    \item Increasing trust in research results through applicability, robustness, fairness, and accountability.
\end{enumerate}

\paragraph{The Structure} of this paper mimics the typical research process in meta-algorithmics. 
We begin by discussing the process of \emph{designing research questions}.
We move on to construct an empirical evaluation for these research questions through high-level \emph{experimental design}.
Next, we investigate \emph{software}, as it is our primary tool: helpful code practices, working with benchmarks and dependencies, and other important considerations for research software.
Lastly, we discuss \emph{interpreting results}, including visualization of results and statistical testing.
All sections include examples of good research practices, as well as common pitfalls to avoid.

We believe this framework is a useful guide to researchers at any level of experience and can serve as a structured introduction to conducting meta-algorithmic research.
Nevertheless, it is not our goal to provide a rule book or a collection of best practices, since specific practices in each field evolve over time.
Indeed, the question of how to do computational research is a core part of our research field itself.
As such, we encourage readers to challenge our recommendations and add their findings to this living document~\footnote{Contributions \& Questions are managed via GitHub: \url{https://github.com/coseal/COSEAL-Best-Practices}}.
Similarly, we commend thoughtful breaking with the practices we present if they do not serve the purpose of the research at hand, as in the spirit of furthering scientific practice, researchers occasionally ``should forget the textbook and simply ask whether he has achieved these higher purposes''~\citep{lessing-69}.

\section{Formulating Research Questions}

Before starting empirical work or experiment design, it is crucial to clarify the objectives of a research project. 
This section makes general suggestions for doing this, specifically by discussing best research practices and formulating research goals.

\paragraph{TL;DR:} Be familiar with the literature on reproducibility and best experimental practices in your field. Decide at the start whether your research project will be confirmatory or exploratory. 
Start with a clear research question or hypothesis.

\subsection{Read Best Practice Literature}
The first step of any research project should be to know the best practices in the field.
Without this background knowledge, even finding interesting research questions can be challenging, let alone designing an experimental evaluation.
Building on a field's established norms, on the other hand, can not only improve a project's outcomes, but also save a lot of research and review effort.

Much research has studied best practices for empirical evaluations and experimental design in and around meta-algorithmics \citep[see, e.g.][]{mockus-bo89a,hooker-or94a,johnson-dimacs02a,mcgeoch-book12a,eggensperger-jair19a,lipton-acm19a}.
Such best practices can contain a range of recommendations to be aware of: from high-level discussions of research directions (see the conversation around metaphor-based meta-heuristics by \cite{aranha-swarm22} and supporting benchmarking results by \cite{vermetten2024largescale}) to general points about the validity of theoretical and empirical results~\citep{hooker-or94a,johnson-dimacs02a,herrmann-icml24} and practical questions of reproducibility and replicability~\citep{mitchell-fat19a,pineau-jmlr21a,bischl-wire23a}, see Table~\ref{tab:best_practices}.
Many domains have their own set of best practices in addition to general recommendations for meta-algorithmic research - some like, e.g., algorithm configuration~\citep{eggensperger-jair19a}, with highly developed standards we can learn from.
Therefore, we recommend reading up on how a specific community has formulated its best practices and where they might differ from others.
This should form a strong basis for formulating interesting research questions that move the field forward and validating them empirically.

We want to add a note on AI assistance at this point since it is relevant here but also for the rest of this paper.
Many researchers from different disciplines use AI tools to help with different research tasks, like summarizing literature, improving writing or creating figures. 
It is certainly worth exploring if AI tools can improve your personal workflows, but especially when it comes to literature research, you should always be able to verify, explain and reproduce the result from an AI tool. 
This means reading the literature first before using AI summaries, thorough corrections of AI-assisted scientific writing and detailed scrutiny of other generated research artifacts like figures.
Also, note the data protection issues associated with many AI services.
When using AI tools, we recommend systems hosted by trusted institutions (many universities now offer such internal services) or systems that allow for some measure of control over data usage rights.

\begin{table}[t!]
    \centering
    \begin{tabularx}{\linewidth}{p{0.2\textwidth} p{0.3\textwidth} X}
        \toprule
        \paragraph{Subject Area} & \paragraph{Best Practices (for)} & \paragraph{Reference(s)} \\
        \midrule
        \multirow{3}{\linewidth}{Empirical Research} & \protect\leavevmode{\color{myblue}Designing Studies} & \cite{herrmann-icml24} \\
        & \protect\leavevmode{\color{mygreen}Comparing Algorithms} & \cite{hooker-or94a,hooker-jh95a,johnson-dimacs02a,mcgeoch-book12a} \\
        \midrule
        \multirow{8}{\linewidth}{General ML} & \protect\leavevmode{\color{myblue}Reproducibility, Writing, Reporting}  & \cite{lipton-acm19a,mitchell-fat19a,pineau-jmlr21a,hofman-arxiv23a,kapoor-arxiv23} \\
        & \protect\leavevmode{\color{mygreen}Comparing Models} & \cite{provost-icml98,demsar-06a} \\
        & \protect\leavevmode{\color{mypink}Visualizing Results} & \cite{hehman-psychsci21,waskom-joss21} \\ 
        \midrule
        \multirow{6}{\linewidth}{AutoML and Optimization} & \protect\leavevmode{\color{myblue}Research and Application} &  \cite{TBB20benchmarking,lindauer-jmlr20a,bischl-wire23a} \\
        & \protect\leavevmode{\color{mygreen}Comparing Algorithms} & \cite{howe-jair02a,eggensperger-jair19a} \\
        \bottomrule
    \end{tabularx}    
    \caption{A non-exhaustive list of literature on best practices in meta-algorithmics. We visually highlight recommendations on best practices in {\color{myblue}General}, for {\color{mypink}Visualization} and {\color{mygreen}Comparisons}.}
    \label{tab:best_practices}
\end{table}

\subsection{Formulate Research Goals}
Broadly speaking, there are two types of empirical research: confirmatory and ex\-plo\-ra\-to\-ry~\citep{schwab-sig20,herrmann-icml24}. 
The goal of confirmatory research is to confirm or disprove an existing theory. 
Exploratory research, on the other hand, aims to gain a basic understanding of an underexplored phenomenon or field where there is, of yet, not enough information to form a well-thought-out hypothesis.
Although this distinction is not always explicitly made in all areas of research that are studied with meta-algorithmic tools, e.g., machine learning~\citep{nakkiran-mleval22}, making this distinction early on will focus the research project and make it more likely that the outcome is useful within the community.

\paragraph{Exploratory Research} means working without a set hypothesis to test. Instead, it is about discovering new questions to answer in confirmatory research~\citep{dietterich-ml90}. However, this does not mean that there should not be clear targets for this kind of research. 
Starting with a research goal that should be based on prior work is important. 
Usually, this goal consists of studying and analyzing an effect or a phenomenon.
The result should be novel insights and research questions that can be tackled in future work.
By its very nature, there is no fixed set of steps to follow in exploratory research, but a community's best practices can guide the process by identifying gaps in understanding and highlighting how to produce meaningful outcomes.
An exploratory work should aim to provide a precise definition of the setting in which it investigates, a discussion of why this setting is both novel and important to study, an investigation of how hard it is to solve, which factors contribute to its solution, and suggestions for future work~\citep{dietterich-ml90}.
The more concrete and thorough the setting, investigation, and suggestions for future work, the more useful this research will be.
Examples of exploratory meta-algorithmic research include works showing that algorithm configuration landscapes tend to be benign, e.g., \cite{pushak-acm22a} or \cite{schneider-ppsn22b}.

\paragraph{Confirmatory Research} should be hypothesis driven. 
That is, one or more hypotheses should be the basis for a new research project.
Such a hypothesis consists of two parts: a research question and a theory about the answer.
It needs to be clear and falsifiable to serve as a basis for new research.
A good research question, like a good theory, is founded on a thorough understanding of all aspects of previous work in the area. 
In fact, many research questions originate as natural follow-up questions to other work. 
Therefore, like in exploratory research, a prerequisite to formulating a good research hypothesis is an extensive base of knowledge of different work in the target area. 
Once this knowledge base is established, there are four important points to keep in mind when forming research hypotheses:

\begin{enumerate}
    \item \textbf{Is the hypothesis grounded in disciplinary standards?} A well-formed hypothesis should build upon established theoretical frameworks and methodological approaches within the field. For interdisciplinary research, ensure the hypothesis respects the epistemological foundations and research standards of all contributing fields rather than violating core principles of any discipline.
    \item \textbf{Is the hypothesis clear and simple?} The practices of all involved fields should be considered and consolidated. Both the research question and corresponding theory should be as short, precise, and simple as possible. Avoid connecting two different theories into one or making the hypothesis imprecise by using words such as ``maybe'', ``possibly'', and ``could''.
    \item \textbf{Is the hypothesis testable?} In most cases, hypotheses should not only be testable in theory, but also in practice. Therefore, also consider whether the amount of available resources is sufficient to answer your research questions in a given study. 
    \item \textbf{Is the hypothesis falsifiable?} 
    The hypothesis should be specific enough that it should be possible to definitively reject it. Avoid grandiose statements targeted at the ``research community at large'' or the distant future.
\end{enumerate}
Formulating concrete research questions upfront can take some time, but they can make experimental design much easier. 
It is good practice to work on a single idea at a time. 
A project's scope may grow over time, but often a single interesting hypothesis is exciting enough for publication and is also easier to communicate to others.

\subsection{Examples \& Pitfalls}

To illustrate our recommendations, we present an example for both exploratory and confirmatory research goals, as well as common pitfalls.

\paragraph{{\color{ForestGreen}Example:} Exploratory Research} We want to focus on investigating how well Bayesian optimization functions in a new real-world application. Specifically, we want to study which solution quality metrics are robust against outliers to avoid destructive results. 

\paragraph{{\color{ForestGreen}Example:} Confirmatory Research} Our leading research question is if early stopping of algorithm configuration using a regret criterion can increase efficiency in terms of trials without degrading performance. 
Our hypothesis is that this approach will result in comparable incumbent performances to full runs on benchmarks where we have already seen early stopping perform well, but that there will still be performance drops on some domains.

\paragraph{{\color{OrangeRed}Pitfall:} No Planning} Not setting goals for a research project is not only inefficient, it can easily lead to improper research practices. Aimless work is more likely to ignore best practices and also more likely to (unwittingly) fall under the umbrella of \emph{method-developing exploratory research} described by \cite{herrmann-icml24}.

\paragraph{{\color{OrangeRed}Pitfall:} Blindly Relying on Best Practices} Especially when conducting novel exploratory research, established best practices can sometimes be limiting. 
Deviations from these norms should always be well-informed and well-thought-out. Norms should be seen as guidelines and be followed in spirit rather than to the letter.

\paragraph{{\color{OrangeRed}Pitfall:} The ``Our Method Is Better'' Hypothesis} A very tempting easy research question and hypothesis to use in confirmatory research takes the form of ``Is our method better?'' - ``Our method outperforms all baselines''. 
This is not a specific question and often not even a testable hypothesis.
``Better'' and ``outperforms'' are very vague terms that can be interpreted in many different ways.
If the goal of the research project is to demonstrate the superiority of a specific method, it should be specified where, how, and why the method works. 

\section{Designing Experimental Evaluations}

With the research questions identified, you can start designing the experimental studies set out to answer these questions.
This chapter is structured by high-level design choices in the experimental pipeline: We start with how to select baseline approaches and the choice of benchmarks to run your experiments.
This is followed up with recommendations on how to set up a fair comparison between different algorithms or learners, and recommendations for reproducible experiments.

\paragraph{TL;DR:}

Always include a simple, well-known baseline approach in your experiments and motivate your benchmark choice by research questions and not the performance of your method. Always configure all approaches in your studies fairly, e.g., by allotting identical configuration budgets, and identify the contributions of each new component you introduce. Set up and report your experiments so that other researchers can effortlessly reproduce your results.

\subsection{Establish Baselines}

Careful calibration of the experimental scope is required to enable researchers to find rigorous and valuable insights.
This initial phase requires thoughtful selection and configuration of various parts of the experimental pipeline, each serving a unique purpose in the overall investigation.
As an important first step, incorporate \emph{relevant baselines}, i.e., simple established algorithms, and \emph{competitive alternatives} to facilitate qualitatively high, comparative analysis while avoiding data overload.
Note that random baselines are often relevant and very simple to run.
Therefore, we recommend including a random baseline in most cases, e.g., random search.

A relevant baseline aids in answering your research questions in detail and is not only there to provide a performance threshold that needs to be beaten.
Further, a relevant baseline should receive the same treatment as the method being studied.
If your own method is subject to extensive tuning, be it manual or automated, the baseline should ideally receive the same attention to enable comparisons on equal footing.
An example of a simple baseline to quantify the benefit of your hyperparameter tuning algorithm in simple terms, is random search with 2x, 4x, 10x, etc. resources~\citep{li-jmlr18a}.
These methods simulate running random search in parallel, which would be the simplest way to assess whether we can improve over simply running random search on more cores (i.e., the most straightforward ``faster'' optimization algorithm).

\subsection{Select Benchmarks}\label{sec:benchmarks}

One key component of the experimental design is the selection of benchmark problems, i.e., test problems and data sets, on which a new system is being evaluated.
The choice of benchmark problem must not be motivated by the desire to show that a novel idea is better than previous ones, but should focus on \emph{generating (positive or negative) evidence for the claims in your project} and to improve understanding of your problem(s) and method(s).
That includes a complete description of the benchmark problems, potential biases, and thorough results reporting (cf.\ below).

An overview of benchmarking guidelines in optimization is provided by \citet{TBB20benchmarking}.
Additionally, early work on the empirical analysis of algorithms can provide guidance on biases that are easily overlooked, e.g., \citet{hooker-jh95a,johnson-dimacs02a}.
In the following, we focus on some practical aspects: the extent and order in which experiments should be run, existing benchmark libraries, and surrogate benchmarking.

\paragraph{Decide on the Number of Benchmarks}
Before choosing one or more specific benchmark problems, we have to decide how many benchmarks to run in the first place.
Running more benchmark problems will, given the results are thoroughly analyzed, allow us to make more substantiated claims about the performance of a system under varying conditions.
However, benchmarking as much as possible for its own sake does not lead to improved understanding of the system to be evaluated and, generally, computational research budgets are finite.
Instead, it is key to select benchmark problems in such a way as to produce a complete picture of within the experimental evaluation, including the strengths and weaknesses of a method.
Furthermore, in some settings such as Hyperparameter optimization (HPO) and AutoML, experiments can be significantly computationally -- and thus economically and ecologically -- expensive.

Benchmark problems should have the \emph{right difficulty} to enable relevant performance analyses, i.e., selected benchmark suites should overall not be too easy or unreasonably hard.
Including some easy and some very difficult problem instances can aid in analyzing performance on these problem classes; however, when only such instances are included in the experimental results, they may be irrelevant or inconclusive at best.
A mix of difficulties is the best approach.
We can also use the problem difficulty to facilitate more efficient benchmarking.
Ordering benchmark problems from easy to hard to solve can help identify bugs earlier and test whether the overall system works before committing significant computational resources to solve more complex problems.
Thus, in case such an ordering of benchmarks is possible, consider using easier benchmarks in a small prototype experiment to refine your design decisions.

Additionally, the number of benchmarks as well as repetitions should be guided by the performance analysis protocol.
Ideally, they allow for robust analysis with respect to statistical tests (cf.~\Cref{sec:post-hoc}).
Note that while the number of repetitions is likely always relevant for post-hoc analysis like statistical testing, the research question may put similar importance on the number of benchmarks, e.g., if the research is on the generalizability of a method. 
Therefore, there are tradeoffs in choosing more benchmarks in favor of less repetitions of each benchmark.
See \citet{hutter-amai10a} for an illustration of such tradeoffs in algorithm design.

\paragraph{Re-use Existing Benchmark Libraries}
For many problem domains, re\-search\-ers have created standardized collections of benchmark problems, for example, ASlib \citep{bischl-aij16a}, BBOB functions \citep{COCOjournal}, HPOBench \citep{eggensperger-neuripsdbt21a}, HPO-B \citep{pineda-neurips21a}, OpenML Benchmarking Suites \citep{bischl-neurips21a}, DACBench \citep{eimer-ijcai21a}) and tools to execute large-scale experiments (e.g., Nevergrad \citep{Nevergrad}, BayesMark \citep{bayesmark}, AutoML benchmark \citep{gijsbers-jmlr24a}, COCO \citep{COCOjournal}, Synetune \citep{salinas2022syne}, IOHprofiler \citep{IOHprofiler} and CARPS \citep{benjamins-arxiv25a}).
Therefore, we can stand on the shoulders of giants, reducing the chances of making implementation mistakes.
Using an established benchmark suite has a multitude of advantages. It relieves the users from selecting test problems and evaluation protocols, including the performance measure(s), number of repetitions, length of runs, etc.
It also allows us to easily compare results between different research groups that evaluated their approach on the same benchmark.

We should take care to select the most \emph{recent version} of a benchmark and document which version we use.
Benchmark suites may be extended over time to include more varieties of problem instances, fix errors in earlier versions, or, in the case of surrogate benchmarks, be more correct and more representative of the real problem.

If current benchmarks are not enough, consider extending and contributing to an existing benchmark instead of introducing a completely new one.
When proposing new test problems, formulate the challenges associated with each of them as explicitly as possible.
For example, the BBOB suite organizes its test problems into groups with shared characteristics and associates research questions with each individual problem \citep{COCOjournal}.

\paragraph{Use Surrogate Benchmarks}
In many situations, running the ``real'' experiments is the factor that contributes the most to the computation time and thus cost.
For example, when evaluating HPO or AutoML systems, training and evaluating a given configuration of your ML pipeline is often the most expensive part \citep[see e.g.,][]{eggensperger-aaai15a,lindauer-jmlr20a}, especially when combined with thorough resampling.

When confronted with a low-dimensional discrete problem space where it is feasible to enumerate the search space, \textit{tabular} benchmarks can be a great starting point.
Here, all possible configurations are evaluated by the benchmark designers beforehand, and the performance data are published to substitute the evaluation pipeline, e.g., training a neural network.
Tabular benchmarks are particularly common in small-scale neural architecture search (NAS) \citep{ying2019bench,mehta-iclr22a,pmlr-v188-pfisterer22a}, but are naturally limited in the expressiveness of their search space due to the curse of dimensionality.

In contrast, surrogate benchmarks replace the evaluation of the ``real'' problem by a surrogate function, which approximates the true problem using a simplified model, e.g., learned by regression.
Thus, they can be viewed as an intermediate step between fully artificial benchmarks (such as BBOB) and real experiments, e.g., of an AutoML system.

A surrogate function can typically be evaluated in fractions of a second on a CPU, rather than multiple hours on a cluster, which has several key benefits.
First of all, the environmental impact of any given experiment is greatly reduced.
The increased experimental speed also allows for faster iterations and testing of more parameter configurations, experimental setups, etc., which is particularly important in the early development of a new system.

The dependencies of surrogate benchmarks are often simpler than those of the real experimental pipeline, as they abstract a lot from the original problem, which in turn reduces the complexity in the implementation significantly.
This improves the transferability of results between different hardware setups, and thus contributes to better reproducibility of experimental data.
The simplified dependencies also remove one source of bugs and errors, especially when running the experiments on HPC clusters, reducing the number of erroneous experimental runs.

Finally, surrogate benchmarks have a democratizing effect.
Not everyone has the enormous HPC resources that are sometimes required, e.g., for running experiments on HPO or AutoML.
The significantly lower compute requirements of surrogate benchmarks enable researchers without access to such resources to perform experiments that were otherwise infeasible for them.

Examples of surrogate benchmarks are YAHPO Gym \citep{pfisterer2022yahpo} for general HPO tasks, as well as some of the scenarios in NAS-Bench-Suite by \citet{mehta-iclr22a} for NAS.

However, we should be careful of drawing very general conclusions from results on surrogate benchmarks.
There are possible inconsistencies with the real task, where the surrogates might exhibit characteristics different from the true system/model. For example, \cite{eggensperger-mlj18a} showed that only by combining performance data from several algorithm configurators for building the surrogate benchmark, they were able to reproduce similar rankings on the surrogate benchmark as on the real one. In some cases, it might be reasonable in case of doubt to also run real benchmarks at the end of an experiment series to verify key results.

\paragraph{Use Synthetic Benchmarks} 
In addition to surrogate benchmarks, some fields of research also offer synthetic benchmarks, i.e., synthetically generated problems or datasets with purposefully created and well-understood challenges.
The main advantage of synthetic benchmarks is that they allow to study specific aspects of a given algorithm or meta-algorithmic framework depending on problem/dataset instance properties: For example, how does an optimizer perform with increasing search space dimensionality (under otherwise unchanged circumstances)?
In addition, they offer potentially even lower computational requirements than surrogate benchmarks.

Synthetic benchmarks are popular in many heuristic optimization domains, e.g., traveling salesperson problems \citep{bossek2019evolving}, combinatorial optimization \citep{Liefooghe2023mocobench}, or (unconstrained) continuous optimization (e.g., BBOB \citep{COCOjournal}).
Furthermore, first synthetic benchmarks for algorithm configuration are available, e.g., SynthACticBench \citep{margraf2025synthacticbench}.

\paragraph{Choose the Right Type of Performance Metric} 
The performance of an algorithm on one instance or dataset can be characterized using different (kinds of) performance metrics, cf.\ Chapter 5 of \cite{TBB20benchmarking}.
A performance metric can measure the quality, time, or robustness of a solution of an algorithm. 
While some benchmarks will specify how to measure performance, generally there are multiple options to choose from.

\emph{Solution quality metrics} relate to the quality of the solution(s) found during the execution of the algorithm.
Examples are the balanced accuracy of a classifier, the solution quality of an optimizer after a fixed budget, or other performance indicators, such as the achieved hypervolume of non-dominated solutions in multi-objective optimization, after termination of the algorithm in question.
Often, solution quality metrics are the easiest to measure, as they do not require much additional information like reference or optimal solutions and can be quickly evaluated.
However, they are also the most sensitive to the experimental setup, e.g., maximal algorithm runtimes / budgets, and cannot always be compared easily across datasets.

\emph{Time-related metrics}, in contrast, measure the time required for an algorithm to reach a specific solution quality. These include, e.g., the expected running time (ERT; \citet{hansen-report09}) prevalent in BBOB and the penalized average running time (PAR)-family of metrics popular in combinatorial optimization scenarios~\citep{hutter-jair09a}.

\emph{Anytime metrics} combine aspects of both solution quality and time-related metrics by tracking how solution quality improves over time during algorithm execution. These metrics are particularly valuable when users have varying time budgets or when the available computation time is unknown in advance. By visualizing or quantifying the quality-time trade-off (e.g., through performance profiles or area under the curve measures), anytime metrics provide a more comprehensive view of algorithm behavior across different time horizons. 

Finally, a \emph{robustness metric} is concerned with the variance in the outcome of multiple runs.

\paragraph{Handle Problems \& Instances with Care}
Most benchmarks will contain several problems, often with different instances (i.e., slight variations of the same underlying problem to avoid overfitting). The exact handling of these instances may depend on the research question and recommendations by the benchmark, but generally we would expect to measure our metrics on problems and instances left out of the optimization process. 
This can be done in the form of cross-validation schemes (see e.g. \citet{hutter-aij14a}) or with leave-one-out evaluation.
We refer to \citet{petelin-gecco25} for an in-depth discussion of leave-one-out evaluation, but recommend a leave-one-problem-out evaluation scheme over leave-one-instance-out options when working across problems as e.g. in algorithm selection. 
We also recommend testing the final results on separate test instances and problems that were not used for the optimization or validation to ensure a reliable estimate of the true performance.

\subsection{Make Design Decisions}

When evaluating a new algorithm, many design decisions regarding the algorithm's (hyper-)parameters, inclusion of different components, etc., have to be made.
Here, we focus on setting the (hyper-)parameters fairly, especially considering comparison and baseline approaches.

\paragraph{Optimize Hyperparameters} 
With the maturity of fields such as algorithm configuration and hyperparameter optimization (HPO), there is a plethora of methods that we can choose from to automatically find well-performing design decisions.
Even though it can be a costly process, this is preferable to manual configuration as hyperparameters can have a significant influence on the empirical results.

First we need to consider which HPO setting is appropriate for our research question. Using the Algorithm Configuration framework~\citep{schede-jair22} to tune our proposed method across a set of tasks to then test the generalizability of this hyperparameter configuration on a left-out task will give us different information about the method's performance than tuning for each task individually. 

Then we need to select an HPO method.
Classical black-box optimization approaches \citep[see e.g.,][]{lopezibanez-orp16a,ansotegui-sat21a,lindauer-jmlr22a} only consider the input-output relationship of the optimization procedure and thus only enable static tuning.
Gray-box optimization methods \citep[see, e.g.,][]{li-jmlr18a,klein-aistats17a,awad-ijcai21a} on the other hand look at multiple such relationships (e.g. input$\mapsto$\{output$_{t=1}$,...,output$_{t=T}$\} relationship at different, discrete times of the target algorithm) to speed up the optimization procedure.
Lastly, white-box optimization approaches \citep[see, e.g,][]{adriaensen-jair22a} continuously monitor the behavior of the target algorithm which ultimately enables any-time configuration.

The choice of which method to employ is highly dependent on the target scenario.
For example, if it is well known that the target algorithm's performance can be approximated well on a subset of the overall available budget, then gray-box methods will be the right choice.
If it is known that the configuration space requires dynamic tuning, white-box approaches should be used.
These examples, however, require domain knowledge, while black-box approaches are reliable choices for tuning if we can not make ready use of such prior information.

\paragraph{Do HPO Fairly} Usually, selecting the right HPO method for the setting is not enough. 
HPO has its own design decisions that need to be made. 
For one, the search space itself needs to be defined.
It is important that it contains existing default values and is broad enough to capture the optimal hyperparameters \citep{anastacio2019exploitation}. 
This is a difficult task without existing domain knowledge, but fortunately, good search spaces can sometimes be determined by referencing established publications (including code bases).
Still, in other scenarios, it can be a challenging endeavor, especially when defaults are determined dynamically based on the input data \citep[see, e.g.,][]{pfisterer2021learning}.

In addition to the search space, we need to select an optimization budget and, depending on the HPO method chosen, additional HPO parameters.
Some prior work suggests including this budget for HPO when calculating overall algorithm budget, though this is not an especially common approach~\citep{jordan-icml20}.
For machine learning, it is also important to choose robust resampling strategies because naive strategies such as the holdout can lead reduced generalization error or overtuning \citep{tschalzev-fmldpr25a,schneider-automl25a}.
Again, these decisions often are not straightforward but can be made by referencing prior work if reasonable.

A crucial factor for a fair evaluation is to standardize all of these decisions across all evaluated methods, including baselines.
Variations for different settings should be well-motivated.
Our goal should be to achieve the best possible performance for every method we execute.
To ensure others can reproduce this part of our experimental pipeline, the HPO process should be documented in detail.

\paragraph{Perform Ablation Studies} The quality of experimentation and the insights gained can be further improved by including ablation experiments that probe the impact of each individual component of a proposed approach, and studying their interaction effects.
As such, ablation studies lend themselves to defining many relevant baselines as they can highlight the impact of individual design choices.
Additionally, ablation studies can be viewed as a crude but informative variant of a configuration study, as ablating individual components can be seen as akin to a grid search over binary hyperparameters that enable/disable parts of the method under study.
Finally, to uncover every aspect of an algorithm's performance, you need to consider possible optimizations concerning feature selection or preprocessing, as well as other design alternatives.

\paragraph{Document Design Decisions} All design decisions should be clearly documented and easy to find. 
This will enable others to reproduce and build upon your current experiments. 
We recommend integrating them into your reporting, either in the paper or alternatively in the supplementary material.
Such transparency on the exact design decisions used is instrumental to focused community progress on the topic at hand and should always be a priority.

\paragraph{Be Aware Of GenAI Experimentation} Generative models are an increasingly important part of our experimental pipeline. 
Evaluating their predictions, however, can be quite involved compared to many other model classes, since generation temperature and the starting prompt play a significant role.
These are serious and important factors to consider when working with generative models.
Prompts are hard to automatically optimize like we would hyperparameters (though attempts have been made to do so~\citep{liu-acm23,wang-iclr23,fernando-iclr24}) yet still deserve a comparable scrutiny. 
It is essential to be familiar with up-to-date best practices and current work in the area of your choice. 
Your strategy around model prompting and evaluation should be documented as well as the rest of your design decisions. 

\subsection{Design Reproducible Experiments}

A study can be reproducible on several levels as \cite{bouthillier-icml19a} describe: its code yield the same outcome every time (Methods Reproducibility), a new implementation of the same method can yield comparable results (Results Reproducibility) and different experimental setups can lead to the same findings (Inferential Reproducibility).
All three of them are important in experimental design since they validate the implementation, the experimental setting and the conclusions.
See also \citet{lopez2021reproducibility} for a discussion of reproducibility in evolutionary computation specifically. 

Ideally, a clear research goal in combination with careful selection of baselines and benchmarks as described above will already set a good basis for Inferential Reproducibility. The following paragraphs add elements that are important for Methods Reproducibility and Results Reproducibility. Methods Reproducibility specifically can be easily tested by re-running a simple toy problem multiple times and we recommend doing so for the best possible experimental basis.


\paragraph{Measure Everything}
The comprehensive measurement of all variables is a fundamental pillar of empirical research.
It ensures reproducibility and transparency of the experiments.
Further, experiments are often expensive, so it is crucial to track everything of interest to avoid rerunning experiments just to track new variables.
Thus, when we refer to \emph{everything}, we refer to a broad spectrum of factors.
In the study of meta-algorithmics, metrics of interest are often not limited to solution quality, but also include runtimes, be it wallclock or used CPU/GPU time.
Generally speaking, it is important to track the chosen and considered solution candidates, CPU/GPU hours, and other compute resources, such as memory.

Taking AutoML as an example, we have various measurements that come into play. For example, all types of machine learning metrics should be captured, such as those computed by scikit-learn \citep{scikit-learn}.
The final pipelines should be serialized to allow post-experimental evaluation of possible additional metrics of interest. Having easy access to such a pipeline can be very helpful, since, for example, a reviewer might ask for a metric they are interested in during a rebuttal.
If storage space permits, predictions on validation and test data, along with the corresponding true labels, should be saved. This enables post-hoc computation of most metrics after the data have already been collected. For classification tasks, the confusion matrix, which provides a comprehensive summary of the algorithm's performance, should also be stored.

An added bonus of measuring everything is that many research artifacts get generated that might be of interest to collaborators or other researchers.
Diligent tracking of interesting data can help lab members avoid having to run expensive data collection themselves.
Thus, it is best to make all the data easily available so that others can use it, optimizing the overall resource utilization.

When advising to measure everything, we have to add a note of caution.
Additional logging can potentially require additional computational resources, which might make fair algorithm comparisons prohibitive.
Thus, runtimes or similar metrics of base algorithms should be logged separately from any other metric being tracked for other analyses.
Packages for resource monitoring exist in several programming languages and are sometimes even built into libraries you may already use.
Job scheduling systems such as SLURM and PBS, frequently employed in high-performance computing (HPC) environments, can also yield valuable data on the resource usage of each execution instance and should be considered when reporting results.
By diligently measuring everything, we ensure that our empirical research in meta-algorithmics is both thorough and replicable and avoid unnecessary waste of resources.

\paragraph{Choose the Right Running Time Measure}

Depending on the goal of the evaluation, it may be required to measure the actual running time on hardware rather than a more abstract measure like function evaluations.
Measuring running time is far from a simple task, since different metrics either eliminate factors like file system speed or general system load or try to account for them. 
The most common metrics include measuring wallclock time, CPU time only or floating point operations per second (FLOPS). 
The selected metric should take all relevant facets of the systems that will be evaluated into account. If a system, for example, creates a lot of load on the file system by design, only measuring CPU time or FLOPS would ignore this extra load and thus not represent the true running time faithfully. 
On the other hand, if the filesystem is slow or under a lot of load due to factors outside of the method's scope, it may be a better choice to opt for CPU time and exclude these filesystem issues from the measurements. 
In some cases, the research question already specifies what kind of time measurement is best, e.g. when the question deals with reducing the number of operations independent of the degree of parallelization. 
In such a case, the wallclock time and CPU time would be heavily influenced by parallelization, while FLOPS allows the focus to shift to total operations.

\paragraph{Use Standardized Data Formats}
The integration of standardized data formats and the consistent annotation of data builds upon the fundamental concept of ``measuring everything''.
Adhering to standardized data formats aids in facilitating comparisons, interpreting results, enabling effective visualizations, and circumvents the necessity of rerunning experiments.
For this reason, many communities agree on and use standardized data formats that make it easy to exchange, access, and read research artifacts.
Such information is often noted down in best practice literature which you should be familiar with.
Standardized formats, in addition to making data collection easier, also help in deciding which data to track. This extends beyond the recording of final results but encourages comprehensive measurement at all experimental stages.
As an example, using standardized data formats enables integrating results in annotated data collections (see the OPTION ontology by \cite{OPTIONtevc} for black-box optimization) or data repositories such as the BBOB data archive at \url{https://coco-platform.org/testsuites/bbob/data-archive.html} or OpenML's~\citep{bischl-patterns25a} collection of datasets, benchmarks, and runs at \url{https://www.openml.org/}.

\paragraph{Build End-to-End Experimental Pipelines}
Creating comprehensive, automated pipelines serves as the backbone of efficient and scalable empirical research. These pipelines should integrate all aspects of the experimental process—from algorithm execution to analysis and reporting—into a cohesive system of interconnected building blocks. 

Your pipeline should reuse logic across different experimental scenarios, such as your primary algorithm, ablation studies, or baselines, ensuring consistency in how experiments are conducted. Each component—data preparation, algorithm execution, logging, result collection, statistical analysis, and visualization—should be modular yet interconnected, allowing experiments to run from start to finish with minimal manual intervention.

Modern tools \citep[see e.g.,][]{biedenkapp-lion18a,tsirigotis-icml18a,sass-realml22a,zoller2023xautoml,fostiropoulos-automlabcd23a,COCOjournal} facilitate this approach by enabling automatic generation of comprehensive reports, including \LaTeX{}-tables of results, statistical test outputs (cf.~\Cref{sec:post-hoc}), and visualizations (cf.~\Cref{sec:plotting}).

This end-to-end pipeline approach offers three key advantages: First, it dramatically reduces human-induced variation and errors that occur during manual handling of intermediate steps. Second, it enhances reproducibility by encoding the entire experimental protocol in software. Third, it creates long-term efficiency gains, as these pipelines can be adapted and reused for future projects. The initial investment in building such pipelines pays dividends through more reliable results and significant time savings over multiple research cycles.

\paragraph{Minimize Sources of Noise and Ensure Statistical Robustness}
Minimizing sources of noise and ensuring statistical robustness are two highly intertwined aspects of empirical research.
Together, they are important aspects of trustworthy scientific results.
The goal is to design experiments such that they maximize the robustness of the generated results, which inherently involves accounting for sources of noise.
Unfortunately, there is no unified framework and notation for defining robustness, especially relating to experimental AI pipelines including optimization, machine learning, and statistical principles.
Still, all state-of-the art concepts have their roots in robust statistics \citep[see e.g.,][]{maronna_robust_2019} founded in the 1960s.
These principles, including robustness against the underlying data distribution, outliers, missing data, and generalization capabilities, should form the fundamental considerations of experimental setups.

An important distinction for our purposes is variance inherent in methods we compare and noise introduced through outside factors.
Noise should be minimized, since it is by definition not relevant to the questions we study. 
Therefore, we recommend using consistent environments for compiling algorithms, which include the same hardware, compiler version, and libraries. In the context of high-performance computing (HPC), ensuring that all runs are conducted on the same partition is crucial. The adoption of containerization wherever possible further aids in the reduction of noise.
Repeating the same run multiple times can reduce noise further.

Variance, on the other hand, has meaning relevant to our research question. 
In contrast to noise, we should thus strive to understand variance.
To do so, we should account for as much variance as we can by varying factors like, e.g., initializations instead of keeping them constant in an attempt to get more consistent results with smaller standard deviations. 
Accounting for more sources of variance within a comparison, in fact, produces a smaller standard error and more reliable comparisons~\citep{bouthillier-mlsys21a}.
Therefore, we recommend minimizing uncontrolled sources of noise as much as possible while at the same time randomizing controllable sources of variance to the highest degree the computational budget will allow.


\subsection{Examples \& Pitfalls}

\paragraph{{\color{ForestGreen} Example:} Simple Baseline Investigation} We want to understand if our novel early stopping mechanism for algorithm configuration actually improves performance compared to standard approaches. To investigate this, we will run the same algorithm configurator with and without early stopping on a set of established benchmarks, measuring both final performance and resource consumption.

\paragraph{{\color{ForestGreen} Example:} Comprehensive Benchmarking Strategy} We aim to thoroughly evaluate our new hyperparameter optimization algorithm against established methods. Our investigation will use both surrogate benchmarks (for rapid, broad testing across many scenarios) and selected real-world problems (to validate surrogate findings). We hypothesize that while our method will show advantages on surrogate benchmarks, some unique characteristics of real-world problems may reveal important limitations.

\paragraph{{\color{OrangeRed} Pitfall:} Blindly Relying on Surrogates} As surrogates offer a low-cost alternative to real-world algorithm runs it is tempting to use them for a large variety of evaluations. However, surrogates can fail to capture all details of a real-world scenario. It could thus very well happen that a shortcoming of a new method is overlooked as surrogates abstract away the root cause of such a shortcoming.

\paragraph{{\color{OrangeRed} Pitfall:} Asymmetric Evaluation} We are investigating a novel hyperparameter optimization method and spend 80\% of our computational budget fine-tuning its components. While this yields strong results in our evaluation, we allocate only minimal resources to baseline methods. This prevents meaningful comparison and risks overstating our method's capabilities, as the baselines weren't given equal opportunity for optimization.

\paragraph{{\color{OrangeRed} Pitfall:} Insufficient Metrics Collection} We run extensive experiments comparing our novel scheduling algorithm against state-of-the-art methods, collecting only the final performance scores. Some time later, we discover that memory usage patterns are crucial for understanding the algorithm's behavior. Without stored measurements, we must rerun all experiments - consuming another 2000 CPU hours that could have been avoided by comprehensive metric logging during the initial runs.

\paragraph{{\color{OrangeRed} Pitfall:} Untested Experimental Setup} We develop a new algorithm configuration method and share our benchmarking framework. When another research group tries to validate our results, they discover missing configuration files and conflicting library dependencies. Our reported performance gains prove unreplicable due to undocumented preprocessing steps. A simple verification run by a co-author would have caught these issues before publication, saving weeks of debugging effort across the community.

\section{Writing Research Software}

Good software is the key to following through on the experimental design of the last section. In the following, we discuss all the elements that need to be considered when implementing algorithms and performing experiments. In addition, we highlight reproducibility aspects that are particularly important for research software and finally discuss how to best utilize resources for experimentation.

\paragraph{TL;DR:} 
Open-source all research artifacts for reproducibility, including code, data, and configurations. Design research code with proper dependency management, quality assurance, and thorough documentation. Start with small prototype experiments, estimate resource requirements, and optimize for efficiency. When using clusters, be mindful of resource allocation, implement checkpointing, and monitor experiments closely.

\subsection{Open-Source Research}
The most important software factor for good experimental practice is making the software accessible in the first place.
This means open-sourcing all software components according to community standards, but it may also make sense to consider standalone publications for software that is used in research.

\paragraph{Make All Artifacts Available}
In order to facilitate open science and research, it is crucial to identify all the artifacts necessary to test and ensure the repeatability and reproducibility of your experiments.
To this end, it is important to make the data, code, as well as all other relevant research artifacts available.
Details about (hyper-)parameter configurations and how to access external data you used in your experiments are crucial for reproducibility.
Any design decision likely has an impact on the final experiments and should therefore be adequately documented.
This documentation itself should also be made available.
Familiarity with the target domain's best practices will prove useful in identifying relevant research artifacts.

\paragraph{Follow Community Standards}
Recently, many communities have started to provide reproducibility checklists.
For example, \citet{pineau-jmlr21a} proposed a reproducibility checklist for machine learning research which can assist as a rough guide when deciding what should be made available and also serves as a reminder to avoid overlooking crucial details.
Beyond such reproducibility checklists, the respective communities have additional recommended standards for reporting and sharing research artifacts. 
For example, in the machine learning community, the so-called model cards \citep[see, e.g., ][]{mitchell-fat19a,crisan-facct22a} are used to enhance transparency by providing detailed, and, importantly, standardized documentation that helps users understand the capabilities, limitations, and appropriate use of machine learning models.
Once all required artifacts have been identified, they should be made available through appropriate platforms that ensure long-term accessibility.
For example, Zenodo\footnote{https://zenodo.org/ is projected to be maintained for the lifetime of the host laboratory CERN, \href{https://help.zenodo.org/guides/nih/element4/}{defined as at least the next twenty years}.} is an open-access repository developed by CERN under the European OpenAIRE program, which allows researchers to share and preserve their research output, including publications, datasets, and software.
Other options include platforms like OpenML~\citep{bischl-patterns25a}, GitHub, and HuggingFace.

\paragraph{Consider Publishing Software}
Software publications serve as vital channels for knowledge dissemination, introducing the wider communities to cutting-edge algorithms, techniques, and tools. The peer review process inherent in most venues helps validate and refine the presented work, ensuring its quality and reliability. These publications also act as formal documentation, creating a lasting record of software developments and methodologies. By providing detailed information, they enable other researchers to reproduce and extend the work, fostering cumulative progress in the field. Moreover, software publications often address specific challenges, offering solutions that can be applied to similar problems across various domains, thus accelerating problem-solving in the broader software ecosystem.
The JMLR Machine Learning Open Source Software (MLOSS) track, the Journal of Open Source Software (JOSS) and the Journal of Statistical Software (JSS), for example, publish short papers on software, and the Evolutionary Computation Journal accepts shorter software articles as well\footnote{\url{https://direct.mit.edu/evco/pages/submission-guidelines\#software}}. This makes your software a first-class citizen that people can explicitly give you credit for through citations, and hopefully provides incentives for continued development. 

\subsection{Design Research Code}
\label{sub-sec:code}
Research code historically has a reputation for being poorly designed and thus difficult to build upon. 
While the expectation should likely not be at a level of production-ready solutions, there are principles that can improve the legibility, structure, and longevity of research code.

\paragraph{Invest in Dependency Management}
Good research code enables others to independently reproduce or replicate your results.
Reproducing results, in turn, requires installing and running the corresponding code.
This can be nontrivial as different platforms or operating system versions can change the availability of packages or how they can be installed.
Therefore, the target audience for the code should be considered: easy installation carries more weight for a benchmark than for an algorithmic improvement paper.
Scripting languages offer environment specifications, such as, e.g., Python's UV~\citep{uv}.
However, these are not infallible and can cause problems across different platforms.
Containers are a reliable alternative to ensure smooth installation.
They are built from recipes that specify many more details, such as the exact operating system on which to build or install the code, and thus ensure that the code is always executed in exactly ``the same environment''.
The most popular options are Docker~\citep{merkel-linux14a} and Singularity~\citep{kurtzer-plos17a}. 
Setting them up requires extra effort, and even container recipes age, but especially when the same functionality is required across platforms, containers are the recommended way of sharing software.

\paragraph{Account for Failure}
It is not unusual for experiments to fail to complete. What is important is to be able to quickly gain insight into why this happens and where the issue lies. 
Therefore, code should contain mechanisms that make it easy to distinguish different types of errors, e.g. between a job failing to launch or the code execution failing, and where they occur. 
Print statements that trigger at important points in the pipeline can be helpful for this (e.g. printing whenever a new function evaluation is scheduled) and can also be used to record important debug information like arguments for a new job. 
Assertions in the code can fulfill similar functions and prevent errors from propagating (e.g. asserting that the last evaluation result is actually a number). Keeping these records for all runs can serve as a valuable resource and save a lot of debugging effort down the line.

\paragraph{Use Code Quality Tools}
Quality code is a major factor in keeping code reproducible and should be ensured as early as possible, as it lowers the probability of surprises, bugs, and costly refactoring later on. 
Moreover, quality code will improve the readability and extensibility, which will allow other researchers to build upon your code and foster collaborations.
Some aspects of code quality can be checked automatically using:
\begin{itemize}
    \item Formatters (e.g., black~\citep{black})
    \item Linters (e.g., ruff~\citep{ruff})
    \item Type checkers (e.g., mypy~\citep{mypy})
    \item Tools for checking compliance with docstring conventions (e.g., pydocstyle~\citep{pydocstyle})
    \item Links to external websites in the documentation
\end{itemize}
Pre-commit~\citep{precommit} loops can run all of these checks on each commit, and thus include them in the project early on. 
Furthermore, unit testing is useful for ensuring code functionality and is therefore recommended for most projects.
Continuous integration / continuous deployment (CI / CD) tools such as Github Actions, CircleCI, and Jenkins can also help to automate the building, testing, and detecting resulting errors after each commit.
We recommend setting up a combination of these tools before writing any code to ensure high code quality standards.
Furthermore, it is worth remembering that scripts to run experiments, gather data, and produce plots are also code and demand the same scrutiny.
AI-based coding tools can help to produce cleaner code as well. 
There are a plethora of different options, from directly interacting with an LLM chatbot interface to IDE integrations such as GitHub Copilot for VSCode~\citep{copilot} and even IDEs fully focused on AI assistance like Cursor~\citep{cursor}.
Integrated solutions are generally using specialized code models and offer greater usability. 
All AI tools should be seen as a complement to the code quality tools above and be used with an awareness that they can produce clean-looking, executable, yet wrong code.
Their output should always be checked before deploying it in a project.

\paragraph{Maintain Documentation}
A big factor in reproducing results will be understanding how this can be done in the first place.
Therefore, proper documentation is essential and should be planned from the start.
Documentation does not only or necessarily refer to a website with an API reference but also includes docstrings, high-quality README documents, and clear instructions for installation and experiment execution.
At the very least, an executable with a thorough README on how to install and use it should be available. 
When using compiled languages, it is crucial to generate build tool files like \texttt{make}/\texttt{cmake} and ``autoconfig'' files to explain conditional compilation. 
A clear and comprehensive README in combination with tool files, installation commands, code structure documentation, docstrings in all methods, and details on how to execute experiments will enable others to reproduce your code later on.

\subsection{Manage Experiments}
Beyond other people interacting with your experiment code, there is also the aspect of using the code to execute experiments in the most efficient way, avoiding excessive re-runs and using only the compute resources required.

\paragraph{Start with a Small Prototype Experiment}
Starting small is often the key to building successful experiments.
Before diving deep into extensive experimental designs, it is recommended to begin with a small prototype experiment. 
Surrogate benchmarks (cf.\ \Cref{sec:benchmarks}) or recorded experimental data, if available, can be invaluable at this stage. 
In cases where such data are not at hand, generating dummy data can serve as an equally effective placeholder. 
This could be as simple as trimming down larger datasets or manually designing small data samples. 
The core aim is not necessarily the precision of the data, but its ability to facilitate swift trial runs that can quickly detect bugs in the pipeline. 
This preliminary stage is crucial to testing the evaluation pipeline holistically. 
This ensures that aspects such as logging, plotting, and resource tracking work efficiently and in tandem. 
As an added advantage, publishing these prototype experiments, in addition to the full set-up, can act as a testament to the replicability of one's research. 
It offers peers a chance to ensure that it seamlessly integrates with their set-ups.
This type of information gathering aligns with the principles of hypothesis-driven research, often providing valuable insights into the assumptions made to form the research questions. 
Depending on the outcome, it can guide the research process backward, prompting a revisit to previous sections, or equipping it with data, laying the groundwork for expansive experiments.
In other words, you should use this stage to verify and validate the design decisions you made before and, if necessary, change the design decisions as needed.
This is also a good opportunity to have one of your co-authors repeat the prototype experiments to ensure your results are repeatable and thus your software setup ready for more extensive experiments.

\paragraph{Perform Regular Backups}
A simple but often overlooked step is to establish a backup strategy.
As soon as you have collected all your results for the first time, start a backup or, better yet, integrate this into your experiment design. 
This is especially important when working on local clusters, which might not guarantee the availability or persistence of workspaces. 

\paragraph{Estimate Resource Requirements}
Understanding and accurately gauging the resource requirements, in particular the runtime requirements, of your experiments is another crucial aspect of experimental design.
Having spent the time to design a small-scale scenario can prove instrumental here.
Analyzing the requirements on a smaller scale is often enough to fairly accurately extrapolate the requirements for full-scale experiments.
In doing so, we should always weigh the potential benefits of running particular experiments against the costs involved.
To gain a better understanding of the overall cost, you should consider various aspects that are affected by the runtime.
For example, you can translate the runtime required for an experiment into CO\textsubscript{2} emissions, energy consumption, monetary costs of running the experiment in a commercial cloud, or allocated CPU / GPU time on a shared local cluster.

\paragraph{Optimize Resource Usage}
Although at first glance it might seem that you have little control over these factors, almost all the design decisions that you face will have some influence on the overall runtime and cost of your experiments.
Take, for example, automated algorithm configuration as a potential application.
Although the overall optimization budget obviously impacts how much resources you use, a more nuanced choice can be made when deciding \emph{how} resources are utilized.
Techniques such as adaptive capping \citep{hutter-jair09a} or multi-fidelity optimization \citep{li-jmlr18a,klein-aistats17a} can be used to quickly discard underperforming solution candidates and thereby avoid wasting resources \citep{eggensperger-jair19a,karapetyan2019,desouza2022}.
This highlights that efficient experimental design should not only consider the overall budget, but also ensure that the available budget that is being spent is used as efficiently as possible.
Having gained a good understanding of the important design decisions, how to verify and test them at a smaller scale, and a solid understanding of the involved resource requirements, we can finally start running our full-scale experiments on the desired benchmark.

\subsection{Adapt Experiments to Clusters}
Full-scale experimentation in meta-algorithmic fields is often done on research clusters. Since these are communal tools with their own load management systems, there are additional considerations for best utilizing them.

\paragraph{Use Resources Mindfully}
Request only the resources necessary for a given job.
In most HPC scheduling systems, this will improve your position in the queue and it will also ensure others can use free resources. 
Jobs can also be optimized for cluster scheduling: Small experiments will probably run fast and are, therefore, scheduled early, but too small experiments usually have negative effects on the scheduling due to the sheer number of jobs submitted. Therefore, it might be a good idea to combine multiple small experiments in a single job. 
This is also dependent on the specific HPC: If there are only huge nodes available, combine as many experiments to fully use each node.
Checkpointing should be done as often as possible when working on an HPC, since even when nothing is wrong in the code, cluster nodes can sometimes fail. For iterative algorithms, it should be possible to save the algorithm state after each iteration. 

\paragraph{Monitor Jobs}
Enable live monitoring of your experiments. Although the experiment might be designed with much care, issues may arise when running experiments on the clusters. It helps a lot to monitor the progress of experiment executions in an appropriate manner to interrupt the execution early when there is a systematic problem. Here again, it also helps to log everything that might be considered useful.

\paragraph{Take Advantage Of Cluster Capabilities} There are common features on HPC clusters that can be very useful for running large-scale experiments. We recommend getting familiar with concepts like job arrays, which allow for starting a whole batch of run variations at once, automatic restarts of failed jobs or sequential execution of multiple pipeline steps. HPC systems are quite powerful and taking advantage of these features can make experimentation a much smoother experience.

\subsection{Examples \& Pitfalls}

\paragraph{{\color{ForestGreen} Example:} Informed Open Source Strategy}
We are working on a benchmarking project that compares several optimization algorithms across benchmarks.
Since we want to make it easy to reproduce these results but work with benchmarks that have conflicting requirements, we choose to create multiple containers which we publish together with our code.
The resulting data can be quite interesting for the community, so we will process our results as well as the resource consumption of each run into .csv files and share them on a code platform like huggingface.
We believe we do not necessarily need to document the established algorithms and benchmarks we use, but focus our efforts on creating clear and simple instructions for running our code through a structured README file, commented runscripts and a documentation page that outlines how algorithms and benchmarks interact.

\paragraph{{\color{ForestGreen} Example:} Improving Code Quality Between Projects}
We want to extend one of our previous projects on evolutionary algorithms where we have not thought as much about our code quality.
Thus we first write test cases for our extension to guarantee we will not accidentally change the original code's functionality.
We also run a few small experiments on an inexpensive problem in order to compare results later.
Then we refactor our code structure, using automated tools to aid in formatting and proper documentation.
After we verify that the tests and experiments still work on the new code base, we use pre-commit loops and GitHub actions to make sure our standards remain high as we extend our method.

\paragraph{{\color{ForestGreen} Example:} Proactive Resource Management}
When investigating a new NAS method, we start working on surrogate benchmarks and then a dummy search space of very small models before moving on to our actual problem setting.
We can use the surrogate benchmarks to test the resource requirements induced by our method and its design decisions.
The dummy search space helps us verify these insights by adding real function evaluations, albeit at a small scale.
With these insights, we have a fairly good idea of what resources to request once we scale to our HPC, improving the rate at which our jobs are scheduled.
If we need to alter design decisions in our experiments, we can immediately adjust the resource requirements and thus our project timeline.

\paragraph{{\color{OrangeRed} Pitfall:} Code as a Last Priority}
It is sometimes tempting to draft experimental code first and and resolve to clean it at some point during the project. 
Not only will this require considerably more effort than to fix inconsistencies immediately, reworking code that has already been used for experiments can potentially lead to functional changes that can make it necessary to rerun everything.
Taking the time to first set up a structure for good research code is likely to be significantly simpler: making explicit choices for installation, setting up a makefile that can take care of automatic formatting and linting and creating a documentation draft that can be filled step by step are simple steps that do not take a lot of time once you know how, but they offer a much-improved workflow when it comes to research code.

\paragraph{{\color{OrangeRed} Pitfall:} Insufficient Installation Testing}
Especially when using requirement specifications or virtual environments for installation, testing if installation from scratch leads to runable code on commonly used systems is essential. 
Locally we often do not install all requirements at the same time, potentially leading to conflicting dependencies package managers cannot resolve. 
Furthermore packages can be limited to specific operating systems and their versions, so it is advisable to test installation on a common Unix system like, e.g., Ubuntu that is used in many HPCs. 
This is crucial when writing software on operating systems that are not as widely used in research, e.g., Windows. 

\paragraph{{\color{OrangeRed} Pitfall:} No Data Management Plan}
It is easy to get lost in experimental data, especially when we log many things.
This can lead to confusion about which experiments have already been executed, where which metrics are recorded and which data should be published in what way.
Doing this post-hoc can be a lot of work and you may discover that you ran an especially costly experiment three times by accident. 
Therefore it is helpful to have a plan upfront on how to handle code, datasets, result data and other artifacts. 
This way you can already set up open-sourcing pipelines at the start of a project and you will be sure of where to find data, how to publish it and what to back up.

\paragraph{{\color{OrangeRed} Pitfall:} Going Big Immediately}
Deploying the first software prototypes immediately on large-scale experiments on an HPC is generally not a good strategy. 
Having to wait on resources only to discover additional errors in the code is frustrating and wasteful of both time and resources.
Gradually moving from smaller experiments to more costly ones ensures that the code is thoroughly tested once more compute resources are required, thus reducing the amount of bug hunting waste.
Ideally, the local setup is as close as possible to the HPC setup as possible (with only the configuration of the job submission being different). Also, many HPC systems provide a test queue that can be used to test the cluster setup.

\paragraph{{\color{OrangeRed} Pitfall:} Only Partial Artifact Release}
Releasing parts of your experimental setup (like partial code or only the resulting model weights) on their own is not equivalent with releasing the full project.
As we discussed, technical details related to versioning or pre-processing can make a big difference for the end result. And to put a result into context, we need to see the full pipeline of how it was created. 
Therefore all code related to the experimental setup, including installation, data loading, runscripts, model code and plotting scripts should be released.

\section{Interpreting Results} \label{sec:interpreting}

After running all experiments, the collected raw data need to be analyzed to extract information about the performance of different approaches, leading to insights about which work best in which scenario.
Here, we consider the following core topics: performance metrics (and where to pay attention when accumulating them), post-hoc analysis, and pitfalls and best practices for results visualization.

\paragraph{TL;DR:}
When interpreting results, keep in mind the type of performance metric (solution quality, time, or robustness), and consciously select how to aggregate the results for presentation (per dataset, ranked for all, etc.).
Statistical tests can help quantify the evidence for research hypotheses, however, always report $p$-values and do not fixate on ``statistical significance'' as the only desired outcome.
Finally, design graphical representations that display the data fairly, including performance over time and variation between repetitions, and that are accessible to everyone and in grayscale printing.

\subsection{Performing Post-Hoc Analysis} \label{sec:post-hoc}

After gathering the experimental data, there is a variety of post-hoc analysis you will likely perform.
We focus here on the main considerations when performing statistical tests.


Hypothesis testing quantifies evidence on whether a null hypothesis -- e.g., along the lines of \textit{``new method does not improve performance''} -- can be rejected based on the collected data.
The output of the hypothesis test is a so-called $p$-value, the probability that the observed data (or more extreme data) was produced assuming that the null hypothesis holds true.
A lower $p$-value indicates a higher probability that a statistical test will reject the null hypothesis.
The right hypothesis test for a given experimental setup can distinguish performance measurements and algorithm rankings that are robust from those due to random fluctuations.
Statistical tests are typically used in confirmatory research, and its usage in exploratory or method-developing research has been dubbed "supposedly confirmatory" \citep{herrmann-icml24}.

Statistical tests should always be \emph{defined together with the experimental setup} and not after the results have already been analyzed to get unbiased results and steer clear from the multiple testing problem.
Applying further statistical tests on already run experiments is also known under the term HARKing (hypothesizing after results known) \citep{kerr1998harking}.
For example, applying a test on subsets of the data after statistical significance was not achieved on the full dataset can easily lead to ``false-positive'' statistical tests: when applying a test with a significance level of $5\%$ on $20$ subsets of data, you would expect one ``significant'' result simply due to random chance.

Note that we use the term ``statistically significant'' very carefully here and want to caution against its (over)use: Community standards for what is considered sufficient statistical evidence evolve with time, and overly focusing on delivering ``statistically significant'' results to some fixed significance level contributes adverse incentives to the experimental and publication process.
Rather, we recommend publishing the achieved $p$-values precisely, and interpret the weight of the evidence in a separate discussion.
For further discussion, we refer to \cite{cockburn2020threats,wasserstein2019moving}.


To decide which statistical tests to use, consider using software such as the Python package \texttt{autorank} \citep{Herbold2020}, which automatically performs several checks on the dataset to decide which tests should be used.
When using such software, make sure to export the report on the statistical tests performed and which tests have actually been applied by it.
For another overview of statistical tests in benchmarking optimization heuristics, see Section 6.3 of \cite{TBB20benchmarking}.





\subsection{Performance Metrics}

Here, we discuss the basic properties of performance metrics and some characteristics to keep in mind when accumulating them in common benchmarking scenarios.

\paragraph{Differentiate Analysis Scenarios}
During the analysis, multiple factors need to be taken into account to derive meaningful results.

Firstly, the problem settings should be differentiated when possible.
For instance, avoid aggregating over easily distinguishable problem properties: e.g., in COCO, aggregation over the search space dimensionality of optimization problems is avoided, as it is known before choosing an optimizer \citep{COCOjournal} and thus performance averaged across low- and high-dimensional problems is of less relevance.
In ML, it is advisable to differentiate the analysis by type of task (e.g., binary vs.\ multi-class vs.\ regression) and use a consistent metric for each scenario, that is, to not aggregate over different metrics.
In some cases, it can be informative to normalize the performance measure values, e.g., relative to a simple baseline, to have a better balance between easy and hard problem instances.
Finally, ensure that all algorithms within one analysis have followed an identical evaluation protocol.
This includes not only training time invested in the final run, but also effort spent for configuration and parameter tuning beforehand.

\paragraph{Aggregate Performance Metrics Mindfully}
When presenting results, it is important not to focus on only one aggregated performance value.
Rather, it is more valuable to provide different descriptive statistics (mean, median, minimum, maximum, first/third quartile, standard deviation, \dots) of the performance instead.
On the one hand, consider performance variation between individual runs (with different random seeds) to evaluate the robustness of an individual method.
On the other hand, variation in algorithm rankings between different datasets/domains should be investigated.
See also, e.g., Chapter 7 of \cite{bartzbeielstein2006}.
Visualizations of the results, for example, in the form of boxplots, violinplots, or simple scatterplots, can greatly improve the expressiveness of the results and the perception by the readers (cf.~Section~\ref{sec:plotting}).

Additionally, it is often more informative to present convergence results, i.e., the performance results after different periods of time during the training and optimization process, rather than the results during a single cut-off point.
While this may not always be possible due to additional overhead in the experimental setup, it enables deeper insights into the behavior of compared algorithms over time and with different budgets.
Graphical visualizations of the convergence, such as the runtime profiles used in BBOB \citep{COCOjournal}, are a good fit to convey these kinds of results.


Generally, with many meta-algorithmic problems, we are dealing with the comparison of multiple algorithms on multiple domains.
We can compare the performance of algorithms in two different ways: (1) For individual test cases (e.g., datasets), we can compare the performance across algorithms, e.g., using rank-based methods, and (2) we can aggregate the average ranking of each algorithm across the multiple datasets.

\begin{itemize}

\item \emph{Approach 1: Compare performance on each domain separately}
When comparing $m$ algorithms on $n$ domains, these results can, for instance, be represented in an $n \times m$ table.
A standard t-test can be used to compare the performance of all algorithms for each domain (each line of the table) after verifying that the results follow a normal distribution (using tests such as the Anderson-Darling test).
As performance data is usually not distributed normally \citep{dewancker2016strategy}, non-parametric tests, such as the Wilcoxon rank sum test, must be used in most cases.
Presenting results using this approach might not give a concrete idea of the competitive performance of each algorithm as it is rare that one algorithm always beats other baselines on all datasets.


\item \emph{Approach 2: Compare performance rankings across domains}
When comparing the performance of multiple algorithms across multiple domains, the results are often not commensurable across the domains.
In such cases, performing a non-parametric test (i.e., ranking test) can give a better idea of relative performance and can allow presenting results in a more concise way than individual performance reports. 
A test like the Friedman test can allow comparing average ranks of algorithms on all domains. 
Creating a critical difference (CD) diagram after a post-hoc test allows us to visually and compactly compare the ranking of multiple algorithms on multiple datasets and assess if the differences in rankings are significant.
A point to consider here is that to statistically compare rankings and evaluate the significance of the results, a relatively large number of runs are needed.
\end{itemize}

\subsection{Visualizing Results} \label{sec:plotting}

Insightful results visualizations make your results easier to understand. The right data visualization can convey a high density of information in a much more accessible way than a table containing the same data.

\paragraph{Choose the Right Chart Type}
A lot of design decisions go into creating comprehensive, clear and honest graphs, starting with the chart type.
Depending on the data, visualization libraries offer a lot of different choices of how to present them.
While there may be a set of available chart types that could be used to convey the right information, there are some general rules that you should follow when selecting a chart type.

Firstly, three-dimensional chart representations can often be misleading, as in almost all cases the graphics are presented in two-dimensional media, such as on screen or paper.
The necessary projection easily warps the perspective and the data is not presented accurately anymore.
Placing multiple graphics or traces behind each other in 3D, or adding 3D effects to any kind of chart that conveys the same information in 2D, should always be avoided.

\begin{figure}[t]
    \centering
    \includegraphics[width=.49\textwidth]{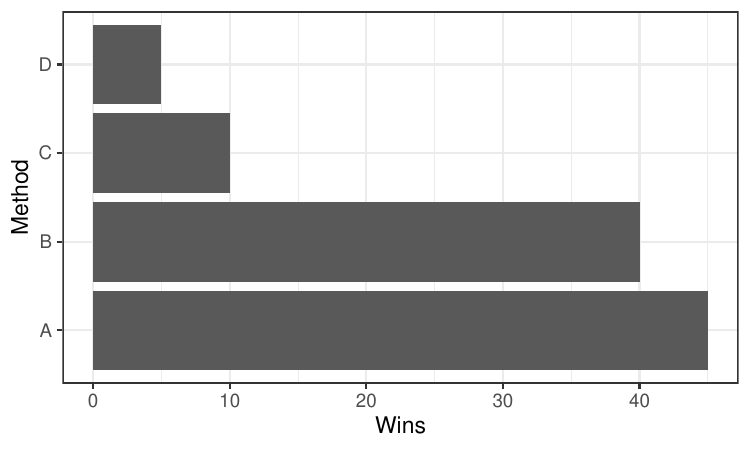}
    \includegraphics[width=.49\textwidth]{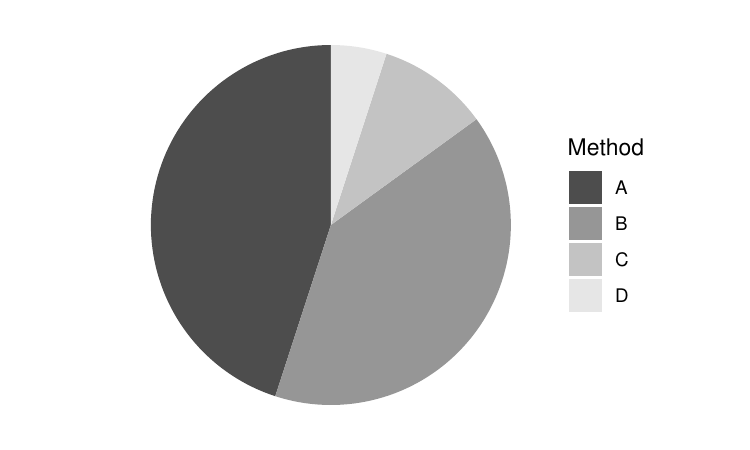}
    \caption{Schematic comparison of a simple barchart vs.\ piechart. Barcharts allow to easily see which approach performs best and quantify performance differences, whereas the piechart is harder to read.
    Additionally, the piechart requires the use of another aesthetic (here: color fill) to distinguish groups.
    This would allow the barchart to include further information or (here) be presented in a more minimalist fashion focusing on the data.}
    \label{fig:chart-comparison}
\end{figure}

Always keep in mind how the data will be used and analyzed by the readers of your research output.
Some kinds of charts are harder to read than others.
For example, pie charts do not allow us to easily read and compare the proportions of each slice.
The same information can in all likelihood be presented in a more readable way in a (stacked) bar chart, as bar lengths are easier to read and intuitively compare than the area/angle of circular sectors.
See Figure~\ref{fig:chart-comparison} for an example.

To better select the type of chart, it is worth looking at the hierarchy of reading characteristics by \citet{datavis}. 

\begin{figure}[t]
    \centering
    \includegraphics[width=.49\textwidth]{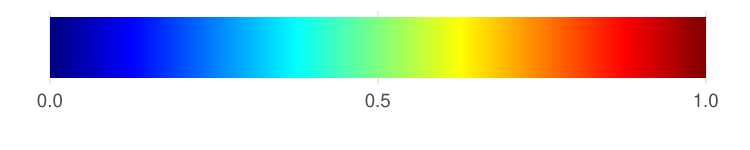}
    \includegraphics[width=.49\textwidth]{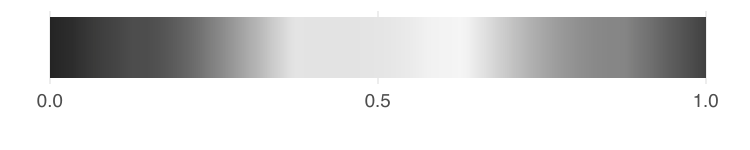}
    \includegraphics[width=.49\textwidth]{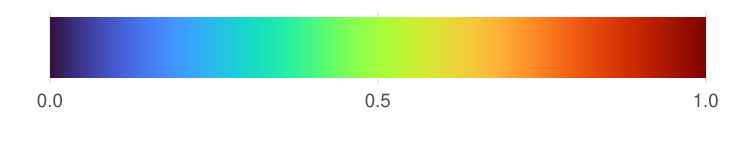}
    \includegraphics[width=.49\textwidth]{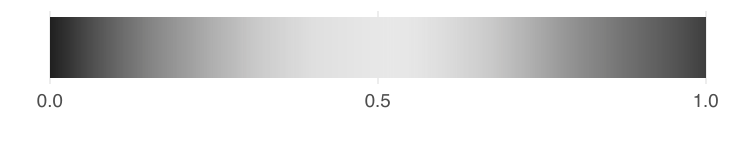}
    \includegraphics[width=.49\textwidth]{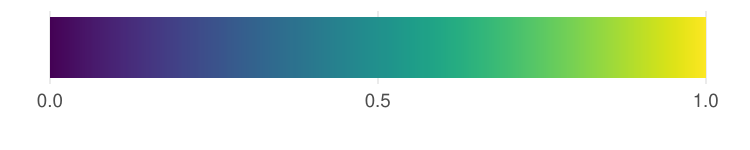}
    \includegraphics[width=.49\textwidth]{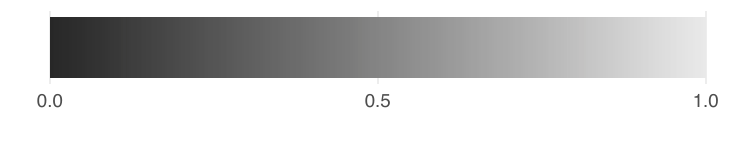}
    \caption{\texttt{jet}, \texttt{turbo} and \texttt{viridis} color scales (top-to-bottom) in color (left) and graytone (right). Of the two rainbow color palettes, \texttt{jet} introduces visual artifacts in forms of brightness jumps, while \texttt{turbo}'s transition between colors is smooth. In contrast to both, \texttt{viridis} is perceptually linear, i.e., brighter colors always correspond to higher values.}
    \label{fig:color-palettes}
\end{figure}

\paragraph{Use Accessible Color Palettes}
When using colors in your charts, pay attention to the ease of identification by people facing color vision deficiencies and use color maps that do not distort color perception \citep{crameri2020misuse}.
Popular and widely available colorblind-friendly palettes are \texttt{viridis} and \texttt{inferno}, among others.
If you want to use ``rainbow'' colors for a wider range of hues, instead of traditional color maps such as \texttt{jet}, consider using a modern variant such as \texttt{turbo} which is accessible to all variants of colorblindness except achromatopsia and, correspondingly, grayscale printing \citep{mikhailov2019}.
Fig.~\ref{fig:color-palettes} illustrates some of these effects.

If you are uncertain, you can use one of various free (online) tools that are available to simulate common colorblindness conditions\footnote{e.g., \url{https://www.color-blindness.com/coblis-color-blindness-simulator/}}.
However, an easy check for color accessibility is to verify that visualizations remain readable when exported or printed in grayscale.

\paragraph{Plot Distributions Instead of Aggregates}
Another contribution a good graphical presentation of results can give, is a closer view into distribution characteristics or uncertainties which go beyond basic summary statistics.

\begin{figure}[t]
    \centering
    \includegraphics[width=.49\textwidth]{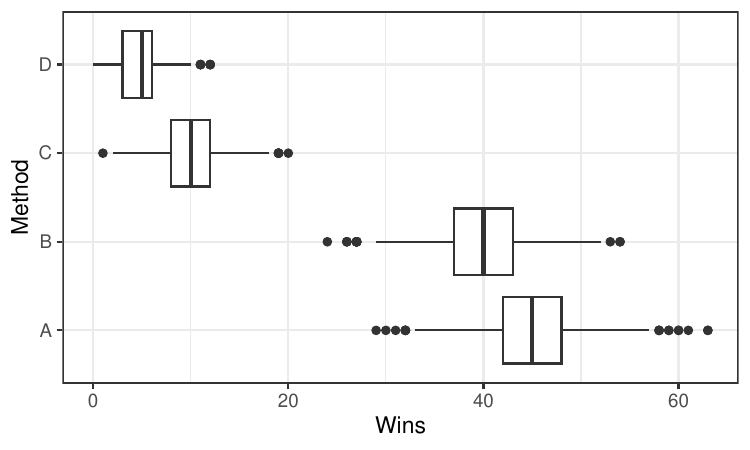}
    \includegraphics[width=.49\textwidth]{figures/barchart.pdf}
    \caption{Schematic comparison of plotting distributed vs.\ aggregated data using a boxplot of individual run data and a barchart of aggregated mean values.
    The boxplot visualization allows us to see that while method A performs better in the median, method B is competitive in some cases as well, while they are both clearly better than methods C and D.}
    \label{fig:distributional}
\end{figure}

Rather than reporting only one averaged metric (such as mean accuracy), boxplots and violion plots can give much better insight into spread and distribution of the underlying individual values.
They can, for example, highlight outliers or distribution characteristics such as multiple modes, skew or variance at a glance.
When comparing across problems, the results per domain should be made comparable by normalization with common reference points (baselines) across domains.
If an oracle is available, the difference between the oracle performance and e.g. a random baseline will be especially informative in such a comparison.
A sensible grouping of the problems also becomes important here: aggregation across easily distinguished characteristics (e.g., classification vs.\ regression, different performance measures, or the dimensionality of an optimization problem) should be avoided and the results should (also) be presented individually per group.
If it would overwhelm the plot to show detailed distributions (e.g., in timeseries, see below), including additional summary statistics such as 5\% and 95\% quantiles or standard errors can still give insights on important distributional characteristics in a readable way. An example is shown in Figure~\ref{fig:distributional}.

\begin{figure}[t]
    \centering
    \includegraphics[width=.49\textwidth]{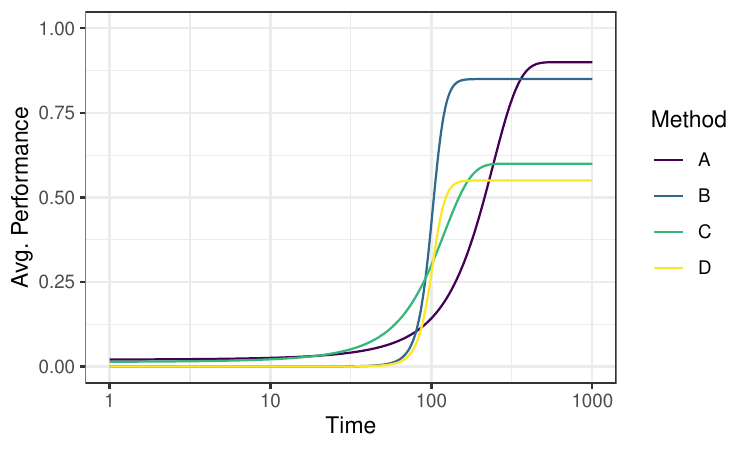}
    \includegraphics[width=.49\textwidth]{figures/barchart.pdf}
    \caption{Schematic comparison of aggregated performance over time vs.\ a fixed point in time.
    The convergence plot shows that, depending on the time allowed, different approaches perform best regarding their average performance.
    For example, Method B outperforms Method A for time budgets in the range of 100-300, while Method A outperforms B after the full time scale of this (hypothetical) experiment.}
    \label{fig:convergence}
\end{figure}

\paragraph{Show Performance Over Time}
Finally, performance data should always be reported at multiple points of computation time or other resource usage, ideally in form of convergence plots.
For example, the COCO platform reports the number of problems solved over time (here: number of function evaluations), giving insights into convergence characteristics.
This also allows others to analyze the trajectories and, e.g., pick an algorithm that has the best results after 1 hour of computation time if that is all you can afford, rather than results after 7 days.
A schematic illustration is given in Figure~\ref{fig:convergence}.

Ideally, performance distributions and performance over time can be shown together in a single graphic.
For example, standard errors or quantile information can be plotted along with median performance by additional lines or color bands.
However, be careful to not introduce unnecessary visual clutter, which can make it hard to make any sense of your visualizations.
Weigh the benefits of including a more accurate representation of the variability of the data vs.\ the readability of your graphic.

\paragraph{Avoid Fully AI-Generated Plots} AI tools have become a popular option to improve presentation in papers. 
They can be helpful with making suggestions such as a suitable color palette, but we caution against workflows that directly input performance data into a model to obtain the corresponding plots.
This is not generally a reproducible process, even with the same model, making it a poor choice.
Furthermore, the way data is aggregated and presented is intimately tied to the research question and the interpretation of the data. 
Therefore, good scientific visualizations, particularly of performance plots, are not easily automated.

\subsection{Examples \& Pitfalls}

\paragraph{{\color{ForestGreen} Example:} Convergence Profiles}
After benchmarking your new hyperparameter optimizer HPO-X, you plot the solution quality of your approach against the running time.
You separately consider different scenarios (numerical, categorical and mixed spaces), and compare against multiple random search baselines (1x, 2x, 5x, and 10x) and the state-of-the-art approach.
You discuss the (relative) performance of HPO-X based on the convergence profiles, keeping the different scenarios and running times in mind.

\paragraph{{\color{ForestGreen} Example:} Statistical Testing}
You decided on a statistical testing protocol beforehand and consider the results of your tests.
In all cases, you report the $p$-values of the tests, and visualize the average ranking and variance of the different methods using critical difference diagrams.
HPO-X may not ``win'' every comparison, but the tests give you a good basis to discuss its strengths and weaknesses.

\paragraph{{\color{OrangeRed} Pitfall:} Out of Control Post-hoc Testing}
After running your experiments, you don't always get the results that you wanted.
Maybe your algorithm did not perform as well as you wished, or the baselines and comparison algorithms are just too good.
Do not try to ``salvage'' the results now by considering an arbitrarily shortened running time, reduced data set or by leaving out state-of-the-art optimizers: Any ``statistically significant'' results received this way are in all likelihood invalidated by the multiple testing.

\paragraph{{\color{OrangeRed} Pitfall:} One Huge Results Table}
There is little less disappointing than reading a paper that has a promising and innovative methodology, transparent and reproducible experimental setup, but chooses to present its results exclusively in tabular form across multiple consecutive pages.
Even with proper statistical tests included, and when highlighting the best algorithm per data set, a few summarizing plots are much better suited to convey the results.
If there is no space for both, consider keeping the figures and publishing the full results table in the appendix or online along your experimental code.

\paragraph{{\color{OrangeRed} Pitfall:} Blindness to Color Blindness}
When selecting color palettes, do not just focus on how appealing certain colors are to you.
Always check accessibility with respect to color blindness, e.g., by checking a grayscale printout, so that other researchers in your community can always interpret your results and figures correctly.




\section{Open Questions}
Research methodology is not fixed, but continuously evolving. As a result, there are elements of meta-algorithmic research practice, we currently cannot formulate best practices for, but believe are important questions in the research community.

\subsection{Impact of Generative AI on Meta-Algorithmics}

Generative Models and their applications have become an important pillar in machine learning research as well as important tools in the sciences and everyday life.
It is not yet entirely clear what this means for the target problems, methods, and workflows in meta-algorithmic research.
Generative AI not only opens new research directions, but has already introduced changes in the research process, like e.g. code generation (see Section~\ref{sub-sec:code}). 
So far, the community has only scratched the surface in investigating how these tools can improve research practice and how we can counter their misuse, as can be seen in the ongoing discussions around LLMs in the review process~\citep{thakkar-corr25}.

\subsection{Data Contamination}
Foundation models today are trained on vast amounts of data, to the degree that for some models, it must be assumed they are trained on all available data at the time of their creation~\citep{magar-acl22,balloccu-eacl24,sainz-conda24}. 
This creates a problem in our usual evaluation workflow with train-test-validation data splits since any validation and test data would have to be kept secret or created after training the model. 

This problem is of course much discussed within the NLP community since LLMs are the best example for models trained on huge amounts of data~\citep{kaplan-corr20,liu-air25}.
Any area using LLMs in their work, however, faces the same issue. 
As an example, let us say we want to use an LLM for algorithm selection. If our benchmarks are openly available, we have to assume any LLM will have been trained on these benchmarks and thus have had access to the test and validation data already.

Currently, we do not know how avoid this issue or even how to measure it well. Iterating on benchmarks fast enough to keep pace with LLM training also is not sustainable. Therefore, it will be an important research topic to pinpoint the influence of data contamination on meta-algorithmic experiments and also to find solutions, e.g. through ideas like unlearning.

\subsection{Long-Term Relevance of Benchmarks}

Benchmarking practice is an important part of meta-algorithmic research. 
The goal is to enable consistent, comprehensive and efficient evaluation of different approaches for a fair comparison.
Often, this means not working on a specific application directly, but to abstract potential downstream tasks for easy execution, e.g. in the form of synthetic functions or by curating a representative set of tasks.
Over time, however, goals in a field may change and what were difficult and interesting target problems are now less relevant or even simple by comparison. 
The question becomes when and how we update benchmarks to stay relevant to the domain they target.
In some domains, this is already a pressing question for current evaluation protocols~\citep{kohli-dmlr24}.

\subsection{Future-Proof Reproducibility Practices}

The central goal of this paper is to encourage broader adoption of good research practice for better reproducibility of results and insights.
It is, however, an open question how we can both spread such practices while continuously improving them. 
In a diverse field like meta-algorithmic research, different sub-fields have their own versions of the practices described here, sometimes for historical reasons, sometimes because of important differences in their problem settings.
Therefore, even spreading current best practices from one domain to another can be a significant effort.
On the other hand, improving how we do research is essential and should be a major goal of the field overall.
Thus, there is a need for better communication and dissemination of research practice across sub-fields in order to produce better quality research in the long term.

\section{Conclusion}
We hope that the experience we summarize in this paper can serve as a guide toward efficient and impactful meta-algorithmic research. 
We expect specifics of what we present here to age and be replaced or supplemented by methods yet to be developed, and hence do not recommend following the letter of this article at any specific point in its lifetime.
The general frame, however, can serve as an entry point to better empirical research for years to come.
We hope it will be enriched by renewed discussion and work on all parts of the pipeline we presented: from better design of research questions to improved practices for experiment setup and more in-depth analysis techniques.
We see this collection of recommendations as a starting point for scientific investigation into research practice, engineering improvements, and knowledge sharing that has the potential to substantially increase the quality and impact of empirical meta-algorithmic research.
We invite such discussions and updates to these practices at \url{https://github.com/coseal/COSEAL-Best-Practices}.
\newrefcontext[sorting=nyt]
\printbibliography[title=References]

@article{adriaensen-jair22a,
  title        = {Automated Dynamic Algorithm Configuration},
  author       = {S. Adriaensen and A. Biedenkapp and G. Shala and N. Awad and T. Eimer and M. Lindauer and F. Hutter},
  year         = 2022,
  journal      = {Journal of Artificial Intelligence Research (JAIR)},
  volume       = 75,
  pages        = {1633--1699},
}

@inproceedings{ansotegui-sat21a,
  title        = {{PyDGGA}: Distributed {GGA} for Automatic Configuration},
  author       = {C. Ans{\'{o}}tegui and J. Pon and M. Sellmann and K. Tierney},
  pages        = {11--20},
  crossref     = {sat21},
}

@inproceedings{awad-ijcai21a,
  title        = {{DEHB}: Evolutionary Hyberband for Scalable, Robust and Efficient {H}yperparameter {O}ptimization},
  author       = {N. Awad and N. Mallik and F. Hutter},
  pages        = {2147--2153},
  crossref     = {ijcai21},
}

@misc{bayesmark,
  title        = {{Bayesmark}: Benchmark framework to easily compare {Bayesian Optimization} methods on real machine learning tasks},
  author       = {R. Turner and D. Eriksson},
  year         = 2019,
  howpublished = {\url{github.com/uber/bayesmark}},
}

@inproceedings{biedenkapp-lion18a,
  title        = {{CAVE}: Configuration Assessment, Visualization and Evaluation},
  author       = {A. Biedenkapp and J. Marben and M. Lindauer and F. Hutter},
  crossref     = {lion18},
}

@article{bischl-aij16a,
  title        = {{ASlib}: A Benchmark Library for Algorithm Selection},
  author       = {B. Bischl and P. Kerschke and L. Kotthoff and M. Lindauer and Y. Malitsky and A. Frech\'{e}tte and H. Hoos and F. Hutter and K. Leyton-Brown and K. Tierney and J. Vanschoren},
  year         = 2016,
  journal      = aij,
  volume       = 237,
  pages        = {41--58},
}

@inproceedings{bischl-neurips21a,
  title        = {{OpenML} Benchmarking Suites},
  author       = {B. Bischl and G. Casalicchio and M. Feurer and F. Hutter and M. Lang and R. Mantovani and J. van Rijn and J. Vanschoren},
  year         = 2021,
  crossref     = {neuripsdbt21},
}

@inproceedings{bouthillier-icml19a,
  title        = {Unreproducible Research is Reproducible},
  author       = {X. Bouthillier and C. Laurent and P. Vincent},
  pages        = {725--734},
  crossref     = {icml19},
}

@inproceedings{bouthillier-mlsys21a,
  title        = {Accounting for Variance in Machine Learning Benchmarks},
  author       = {X. Bouthillier and P. Delaunay and M. Bronzi and A. Trofimov and B. Nichyporuk and J. Szeto and N. Mohammadi Sepahvand and E. Raff and K. Madan and V. Voleti and S. Ebrahimi Kahou and V. Michalski and T. Arbel and C. Pal and G. Varoquaux and P. Vincent},
  pages        = {747--769},
  crossref     = {mlsys21},
}

@article{demsar-06a,
  title        = {Statistical Comparisons of Classifiers over Multiple Data Sets},
  author       = {J. Demšar},
  year         = 2006,
  journal      = jmlr,
  volume       = 7,
  pages        = {1--30},
}

@inproceedings{eggensperger-aaai15a,
  title        = {Efficient Benchmarking of Hyperparameter Optimizers via Surrogates},
  author       = {K. Eggensperger and F. Hutter and H. Hoos and K. Leyton-Brown},
  pages        = {1114--1120},
  crossref     = {aaai15},
}

@article{eggensperger-jair19a,
  title        = {Pitfalls and Best Practices in Algorithm Configuration},
  author       = {K. Eggensperger and M. Lindauer and F. Hutter},
  year         = 2019,
  journal      = jair,
  pages        = {861--893},
}

@article{eggensperger-mlj18a,
  title        = {Efficient benchmarking of algorithm configurators via model-based surrogates},
  author       = {K. Eggensperger and M. Lindauer and H. Hoos and F. Hutter and K. Leyton{-}Brown},
  year         = 2018,
  journal      = {Machine Learning},
  volume       = 107,
  number       = 1,
  pages        = {15--41},
}

@inproceedings{eggensperger-neuripsdbt21a,
  title        = {{HPOBench}: A Collection of Reproducible Multi-Fidelity Benchmark Problems for {HPO}},
  author       = {K. Eggensperger and P. M{\"u}ller and N. Mallik and M. Feurer and R. Sass and A. Klein and N. Awad and M. Lindauer and F. Hutter},
  crossref     = {neuripsdbt21},
}

@inproceedings{eimer-ijcai21a,
  title        = {{DACB}ench: {A} Benchmark Library for Dynamic Algorithm Configuration},
  author       = {T. Eimer and A. Biedenkapp and M. Reimer and S. Adriaensen and F. Hutter and M. Lindauer},
  year         = 2021,
  publisher    = {ijcai.org},
  pages        = {1668--1674},
  crossref     = {ijcai21},
}

@article{hooker-jh95a,
  title        = {Testing heuristics: We have it all wrong},
  author       = {J. Hooker},
  year         = 1995,
  journal      = {Journal of Heuristics},
  pages        = {33--42},
}

@article{hooker-or94a,
  title        = {Needed: An empirical science of algorithms},
  author       = {J. Hooker},
  year         = 1994,
  journal      = {Operations research},
  publisher    = {{INFORMS}},
  volume       = 42,
  number       = 2,
  pages        = {201--212},
}

@article{hutter-aij14a,
  title        = {Algorithm runtime prediction: Methods and evaluation},
  author       = {F. Hutter and L. Xu and H. Hoos and K. Leyton-Brown},
  year         = 2014,
  journal      = aij,
  volume       = 206,
  pages        = {79--111},
}

@article{hutter-amai10a,
  title        = {Tradeoffs in the Empirical Evaluation of Competing Algorithm Designs},
  author       = {F. Hutter and H. Hoos and K. Leyton-Brown},
  year         = 2010,
  journal      = {Annals of Mathematics and Artificial Intelligenc (AMAI), Special Issue on Learning and Intelligent Optimization},
  volume       = 60,
  number       = 1,
  pages        = {65--89},
}

@article{hutter-jair09a,
  title        = {Param{ILS}: An Automatic Algorithm Configuration Framework},
  author       = {F. Hutter and H. Hoos and K. Leyton-Brown and T. St{\"u}tzle},
  year         = 2009,
  journal      = jair,
  volume       = 36,
  pages        = {267--306},
}

@article{johnson-dimacs02a,
  title        = {A theoretician’s guide to the experimental analysis of algorithms},
  author       = {D. Johnson},
  year         = 2002,
  journal      = {Proceedings of the 5th and 6th DIMACS implementation challenges},
  volume       = 59,
  pages        = {215--250},
}

@inproceedings{klein-aistats17a,
  title        = {Fast {Bayesian} Optimization of Machine Learning Hyperparameters on Large Datasets},
  author       = {A. Klein and S. Falkner and S. Bartels and P. Hennig and F. Hutter},
  year         = 2017,
  crossref     = {aistats17},
}

@article{kurtzer-plos17a,
  title        = {Singularity: Scientific containers for mobility of compute.},
  author       = {G. Kurtzer and V. Sochat and M. Bauer},
  year         = 2017,
  journal      = {PloS one},
  publisher    = plos,
  volume       = 12,
  number       = 5,
}

@article{li-jmlr18a,
  title        = {Hyperband: A Novel Bandit-Based Approach to {H}yperparameter {O}ptimization},
  author       = {L. Li and K. Jamieson and G. DeSalvo and A. Rostamizadeh and A. Talwalkar},
  year         = 2018,
  journal      = jmlr,
  volume       = 18,
  number       = 185,
  pages        = {1--52},
}

@article{lindauer-jmlr20a,
  title        = {Best practices for scientific research on {N}eural {A}rchitecture {S}earch},
  author       = {M. Lindauer and F. Hutter},
  year         = 2020,
  journal      = {Journal of Machine Learning Research},
  volume       = 21,
  number       = 243,
  pages        = {1--18},
}

@article{lindauer-jmlr22a,
  title        = {{SMAC3}: A Versatile Bayesian Optimization Package for {H}yperparameter {O}ptimization},
  author       = {M. Lindauer and K. Eggensperger and M. Feurer and A. Biedenkapp and D. Deng and C. Benjamins and T. Ruhkopf and R. Sass and F. Hutter},
  year         = 2022,
  journal      = jmlr,
  volume       = 23,
  number       = 54,
  pages        = {1--9},
}

@article{lopezibanez-orp16a,
  title        = {The irace package: Iterated racing for automatic algorithm configuration},
  author       = {M. L{\'{o}}pez{-}Ib{\'{a}}{\~{n}}ez and J. Dubois-Lacoste and L. Perez Caceres and M. Birattari and T. St{\"{u}}tzle},
  year         = 2016,
  journal      = {Operations Research Perspectives},
  volume       = 3,
  pages        = {43--58},
}

@book{mcgeoch-book12a,
  title        = {A Guide to Experimental Algorithmics},
  author       = {C. McGeoch},
  year         = 2012,
  publisher    = {Cambridge University Press},
}

@inproceedings{mehta-iclr22a,
  title        = {{NAS-Bench-Suite}: {NAS} Evaluation is (Now) Surprisingly Easy},
  author       = {Y. Mehta and C. White and A. Zela and A. Krishnakumar and G. Zabergja and S. Moradian and M. Safari and K. Yu and F. Hutter},
  crossref     = {iclr22},
}

@article{merkel-linux14a,
  title        = {Docker: lightweight linux containers for consistent development and deployment},
  author       = {D. Merkel},
  year         = 2014,
  journal      = {Linux journal},
  volume       = 2014,
  number       = 239,
}

@book{mockus-bo89a,
  title        = {{Bayesian} Approach to Global Optimization. Theory and Applications},
  author       = {J. Mockus},
  year         = 1989,
  publisher    = {Kluwer Academic Publishers},
  editor       = {M. Hazewinkel},
}

@article{pineau-jmlr21a,
  title        = {Improving Reproducibility in Machine Learning Research (A Report from the {NeurIPS 2019} Reproducibility Program)},
  author       = {J. Pineau and P. Vincent-Lamarre and K. Sinha and V. Lariviere and A. Beygelzimer and F. d'Alche-Buc and E. Fox and H. Larochelle},
  year         = 2021,
  journal      = jmlr,
  volume       = 22,
  number       = 164,
  pages        = {1--20},
}

@inproceedings{pineda-neurips21a,
  title        = {{HPO-B}: A Large-Scale Reproducible Benchmark for Black-Box {HPO} based on {OpenML}},
  author       = {S. Pineda and H. Jomaa and M. Wistuba and J. Grabocka},
  year         = 2021,
  crossref     = {neuripsdbt21},
}

@article{pushak-acm22a,
  title        = {AutoML Loss Landscapes},
  author       = {Y. Pushak and H. Hoos},
  year         = 2022,
  journal      = {ACM Transactions on Evolutionary Learning and Optimization},
  publisher    = {ACM},
  address      = {New York, NY, USA},
  volume       = 2,
  number       = 3,
  pages        = {1--30},
}

@inproceedings{sass-realml22a,
  title        = {DeepCAVE: An Interactive Analysis Tool for Automated Machine Learning},
  author       = {R. Sass and E. Bergman and A. Biedenkapp and F. Hutter and M. Lindauer},
  year         = 2022,
  crossref     = {realml22},
}

@article{scikit-learn,
  title        = {Scikit-learn: Machine Learning in {P}ython},
  author       = {F. Pedregosa and G. Varoquaux and A. Gramfort and V. Michel and B. Thirion and O. Grisel and M. Blondel and P. Prettenhofer and R. Weiss and V. Dubourg and J. Vanderplas and A. Passos and D. Cournapeau and M. Brucher and M. Perrot and E. Duchesnay},
  year         = 2011,
  journal      = jmlr,
  volume       = 12,
  pages        = {2825--2830},
}

@article{bischl-wire23a,
  title={Hyperparameter optimization: Foundations, algorithms, best practices, and open challenges},
  author={Bischl, Bernd and Binder, Martin and Lang, Michel and Pielok, Tobias and Richter, Jakob and Coors, Stefan and Thomas, Janek and Ullmann, Theresa and Becker, Marc and Boulesteix, Anne-Laure and others},
  journal={Wiley Interdisciplinary Reviews: Data Mining and Knowledge Discovery},
  volume={13},
  number={2},
  year={2023},
  publisher={Wiley Online Library},
  doi={https://doi.org/10.1002/widm.1484}
}

@article{cockburn2020threats,
  title={Threats of a replication crisis in empirical computer science},
  author={Cockburn, Andy and Dragicevic, Pierre and Besan{\c{c}}on, Lonni and Gutwin, Carl},
  journal={Communications of the ACM},
  volume={63},
  number={8},
  pages={70--79},
  year={2020},
  publisher={ACM New York, NY, USA}
}

@article{wasserstein2019moving,
    author = {Ronald L. Wasserstein, Allen L. Schirm and Nicole A. Lazar},
    title = {Moving to a World Beyond “p < 0.05”},
    journal = {The American Statistician},
    volume = {73},
    number = {sup1},
    pages = {1--19},
    year = {2019},
    publisher = {ASA Website},
    doi = {10.1080/00031305.2019.1583913}
}

@book{bartzbeielstein2006,
    author = {Thomas Bartz-Beielstein},
    title = {Experimental Research in Evolutionary Computation - The New Experimentalism},
    year = {2006},
    publisher={Springer}
}

@article{desouza2022,
title = {Capping methods for the automatic configuration of optimization algorithms},
journal = {Computers \& Operations Research},
volume = {139},
pages = {105615},
year = {2022},
doi = {https://doi.org/10.1016/j.cor.2021.105615},
author = {Marcelo {de Souza} and Marcus Ritt and Manuel López-Ibáñez}
}

@InProceedings{karapetyan2019,
author="Karapetyan, Daniel
and Parkes, Andrew J.
and St{\"u}tzle, Thomas",
editor="Battiti, Roberto
and Brunato, Mauro
and Kotsireas, Ilias
and Pardalos, Panos M.",
title="Algorithm Configuration: Learning Policies for the Quick Termination of Poor Performers",
booktitle="Learning and Intelligent Optimization",
year="2019",
publisher="Springer International Publishing",
address="Cham",
pages="220--224",
isbn="978-3-030-05348-2"
}

@ARTICLE{OPTIONtevc,
  author={Kostovska, Ana and Vermetten, Diederick and Doerr, Carola and Džeroski, Sašo and Panče Panov and Eftimov, Tome},
  journal={IEEE Transactions on Evolutionary Computation}, 
  title={OPTION: OPTImization Algorithm Benchmarking ONtology}, 
  year={2022},
  doi={10.1109/TEVC.2022.3232844}, 
note = {To appear. Free version available at \url{https://arxiv.org/abs/2211.11332}}}

@article{TBB20benchmarking,
  author    = {Thomas Bartz{-}Beielstein and
               Carola Doerr and
               Jakob Bossek and
               Sowmya Chandrasekaran and
               Tome Eftimov and
               Andreas Fischbach and
               Pascal Kerschke and
               Manuel L{\'{o}}pez{-}Ib{\'{a}}{\~{n}}ez and
               Katherine M. Malan and
               Jason H. Moore and
               Boris Naujoks and
               Patryk Orzechowski and
               Vanessa Volz and
               Markus Wagner and
               Thomas Weise},
  title     = {Benchmarking in Optimization: Best Practice and Open Issues},
  journal   = {CoRR},
  volume    = {abs/2007.03488},
  year      = {2020},
  url       = {https://arxiv.org/abs/2007.03488},
  archivePrefix = {arXiv},
  eprint    = {2007.03488}
}

@article{COCOjournal,
author = {Nikolaus Hansen and Anne Auger and Raymond Ros and Olaf Mersmann and Tea Tu{\v s}ar and Dimo Brockhoff},
title = {{COCO:} a platform for comparing continuous optimizers in a black-box setting},
journal = {Optimization Methods and Software},
pages = {1-31},
year  = {2020},
publisher = {Taylor & Francis},
doi = {10.1080/10556788.2020.1808977},
URL = {https://doi.org/10.1080/10556788.2020.1808977},
eprint = {https://doi.org/10.1080/10556788.2020.1808977}
}

@software{Nevergrad,
    author = {J{\'{e}}r{\'{e}}my Rapin and Olivier Teytaud},
    title = {{Nevergrad - A gradient-free optimization platform}},
    year = {2018},
    publisher = {GitHub},
    journal = {GitHub repository},
    howpublished = {\url{https://GitHub.com/FacebookResearch/Nevergrad}}
}

@article{schede-jair22,
  author    = {Elias Schede and
               Jasmin Brandt and
               Alexander Tornede and
               Marcel Wever and
               Viktor Bengs and
               Eyke H{\"{u}}llermeier and
               Kevin Tierney},
  title     = {A Survey of Methods for Automated Algorithm Configuration},
  journal   = {J. Artif. Intell. Res.},
  volume    = {75},
  pages     = {425--487},
  year      = {2022},
  url       = {https://doi.org/10.1613/jair.1.13676},
  doi       = {10.1613/jair.1.13676},
  bibsource = {dblp computer science bibliography, https://dblp.org}
}

@article{datavis,
 ISSN = {00359238},
 URL = {http://www.jstor.org/stable/2981473},
 author = {William S. Cleveland and Robert McGill},
 journal = {Journal of the Royal Statistical Society. Series A (General)},
 number = {3},
 pages = {192--229},
 publisher = {[Royal Statistical Society, Wiley]},
 title = {Graphical Perception: The Visual Decoding of Quantitative Information on Graphical Displays of Data},
 urldate = {2023-04-07},
 volume = {150},
 year = {1987}
}

@online{mikhailov2019,
	month = {8},
	title = {{Turbo, An Improved Rainbow Colormap for Visualization}},
	year = {2019},
    author = {Anton Mikhailov},
	url = {https://ai.googleblog.com/2019/08/turbo-improved-rainbow-colormap-for.html},
}

@article{crameri2020misuse,
  title={The misuse of colour in science communication},
  author={Crameri, Fabio and Shephard, Grace E and Heron, Philip J},
  journal={Nature communications},
  volume={11},
  number={1},
  pages={5444},
  year={2020},
  publisher={Nature Publishing Group UK London}
}

@article{lipton-acm19a,
  title={Troubling Trends in Machine Learning Scholarship: Some ML papers suffer from flaws that could mislead the public and stymie future research.},
  author={Lipton, Zachary C and Steinhardt, Jacob},
  journal={Queue},
  volume={17},
  number={1},
  pages={45--77},
  year={2019},
  publisher={ACM New York, NY, USA}
}

@inproceedings{provost-icml98,
author = {F. Provost and T. Fawcett and R. Kohavi},
title = {The Case against Accuracy Estimation for Comparing Induction Algorithms},
crossref = {icml98}
}

@proceedings{icml98,
  title        = {Proceedings of the 15th International Conference on Machine Learning ({ICML}'98)},
  year         = 1998,
  booktitle    = {Proceedings of the Sixteenth International Conference on Machine Learning ({ICML}'98)},
  publisher    = kaufmann,
  editor       = {J. Shavlik},
}

@article{hofman-arxiv23a,
      title   = {Pre-registration for Predictive Modeling}, 
      author  = {J. Hofman and A. Chatzimparmpas and A. Sharma and D. Watts and J. Hullman},
      year    = 2023,
      journal = {arXiv: 2311.18807 {[cs.LG]}}
}

@article{howe-jair02a,
  title = {A critical assessment of benchmark comparison in planning},
  author = {A. Howe and E. Dahlmann},
  volume = {17},
  year = 2002,
  editor =  {M. Pollack},
  journal = jair
}

@article{kapoor-arxiv23,
      title={REFORMS: Reporting Standards for Machine Learning Based Science}, 
      author={Sayash Kapoor and Emily Cantrell and Kenny Peng and Thanh Hien Pham and Christopher A. Bail and Odd Erik Gundersen and Jake M. Hofman and Jessica Hullman and Michael A. Lones and Momin M. Malik and Priyanka Nanayakkara and Russell A. Poldrack and Inioluwa Deborah Raji and Michael Roberts and Matthew J. Salganik and Marta Serra-Garcia and Brandon M. Stewart and Gilles Vandewiele and Arvind Narayanan},
      year={2023},
      journal = {arXiv: 2308.07832 {[cs.LG]}}
}

@article{waskom-joss21, 
    year = {2021}, 
    publisher = {The Open Journal}, 
    volume = {6}, 
    number = {60}, 
    pages = {3021}, 
    author = {Michael L. Waskom}, 
    title = {Seaborn: statistical data visualization}, 
    journal = {Journal of Open Source Software} 
}

@article{hehman-psychsci21,
    author = {E. Hehman and S. Xie},
    title = {Doing Better Data Visualization},
    journal = {Advances in Methods and Practices in Psychological Science},
    year = 2021,
    volume = 4
}

@inproceedings{salinas2022syne,
  title={Syne tune: A library for large scale hyperparameter tuning and reproducible research},
  author={Salinas, David and Seeger, Matthias and Klein, Aaron and Perrone, Valerio and Wistuba, Martin and Archambeau, Cedric},
  booktitle={International Conference on Automated Machine Learning},
  pages={16--1},
  year={2022},
  organization={PMLR}
}

@inproceedings{pfisterer2022yahpo,
  title={Yahpo gym-an efficient multi-objective multi-fidelity benchmark for hyperparameter optimization},
  author={Pfisterer, Florian and Schneider, Lennart and Moosbauer, Julia and Binder, Martin and Bischl, Bernd},
  booktitle={International Conference on Automated Machine Learning},
  pages={3--1},
  year={2022},
  organization={PMLR}
}

@article{Kaur22, 
author = {Kaur, Davinder and Uslu, Suleyman and Rittichier, Kaley J. and Durresi, Arjan}, 
title = {Trustworthy Artificial Intelligence: A Review}, year = {2022}, 
issue_date = {March 2023}, publisher = {Association for Computing Machinery}, 
address = {New York, NY, USA}, volume = {55}, number = {2}, issn = {0360-0300}, 
url = {https://doi.org/10.1145/3491209}, 
doi = {10.1145/3491209}, 
journal = {ACM Comput. Surv.}, 
month = {01}, 
articleno = {39}, 
numpages = {38} }

@article{TAILORRoadmap22,
    title={{Strategic Research and Innovation Roadmap of trustworthy AI}},
    author={TAILOR}, 
    pages={1--57},
    journal = {https://tailor-network.eu/research-overview/strategic-research-and-innovation-roadmap/},
    year = {2022},
}

@article{EUExpert19,
    title={{Ethics guidelines for trustworthy AI}},
    author={{European Commission}}, 
    pages={1--41},
    journal = {https://digital-strategy.ec.europa.eu/en/library/ethics-guidelines-trustworthy-ai},
    year = {2019},
}

@book{maronna_robust_2019,
	series = {Wiley {Series} in {Probability} and {Statistics}},
	title = {Robust {Statistics}: {Theory} and {Methods} (with {R})},
	isbn = {978-1-119-21468-7},
	url = {https://books.google.de/books?id=K5RxDwAAQBAJ},
	publisher = {Wiley},
	author = {Maronna, R.A. and Martin, R.D. and Yohai, V.J. and Salibián-Barrera, M.},
	year = {2019},
	lccn = {2018033202},
}

@inproceedings{fostiropoulos-automlabcd23a,
  title={ABLATOR: Robust Horizontal-Scaling of Machine Learning Ablation Experiments},
  author={Fostiropoulos, I. and Itti, L.},
  booktitle={AutoML Conference 2023 (ABCD Track)},
  year={2023}
}

@inproceedings{tsirigotis-icml18a,
  title={Oríon: Experiment version control for efficient hyperparameter optimization},
  author={Tsirigotis, C. and Bouthillier, X. and Corneau-Tremblay, F. and Henderson, P. and Askari, R. and Lavoie-Marchildon, S. and Deleu, T. and Suhubdy, D. and Noukhovitch, M. and Bastien, F. and Pascal, L.},
  year={2018},
 booktitle={RML Workshop@ICML'18}
}

@inproceedings{ying2019bench,
  title={Nas-bench-101: Towards reproducible neural architecture search},
  author={Ying, Chris and Klein, Aaron and Christiansen, Eric and Real, Esteban and Murphy, Kevin and Hutter, Frank},
  booktitle={International conference on machine learning},
  pages={7105--7114},
  year={2019},
  organization={PMLR}
}

@article{Herbold2020,
  doi = {10.21105/joss.02173},
  url = {https://doi.org/10.21105/joss.02173},
  year = {2020},
  publisher = {The Open Journal},
  volume = {5},
  number = {48},
  pages = {2173},
  author = {Steffen Herbold},
  title = {Autorank: A Python package for automated ranking of classifiers},
  journal = {Journal of Open Source Software}
}

@article{aranha-swarm22,
  author       = {C. Aranha and
                  C. Camacho{-}Villal{\'{o}}n and
                  F. Campelo and
                  M. Dorigo and
                  R. Ruiz and
                  M. Sevaux and
                  K. S{\"{o}}rensen and
                  T. St{\"{u}}tzle},
  title        = {Metaphor-based metaheuristics, a call for action: the elephant in
                  the room},
  journal      = {Swarm Intell.},
  volume       = {16},
  number       = {1},
  pages        = {1--6},
  year         = {2022},
  doi          = {10.1007/S11721-021-00202-9},
}

@article{dietterich-ml90,
  author       = {T. Dietterich},
  title        = {Exploratory Research in Machine Learning},
  journal      = {Machine Learning},
  volume       = {5},
  pages        = {5--9},
  year         = {1990},
  url          = {https://doi.org/10.1007/BF00115892},
  doi          = {10.1007/BF00115892},
}

@article{schwab-sig20,
  author       = {S. Schwab and L. Held},
  title        = {Different worlds Confirmatory versus exploratory research},
  journal      = {Significance},
  volume       = {17},
  pages        = {8--9},
  year         = {2020},
  url          = {https://doi.org/10.1111/1740-9713.01369},
  doi          = {10.1111/1740-9713.01369},
}

@inproceedings{herrmann-icml24,
  author       = {M. Herrmann and
                  F. Lange and
                  K. Eggensperger and
                  G. Casalicchio and
                  M. Wever and
                  M. Feurer and
                  D. R{\"{u}}gamer and
                  E. H{\"{u}}llermeier and
                  A. Boulesteix and
                  B. Bischl},
  title        = {Position: Why We Must Rethink Empirical Research in Machine Learning},
  booktitle={International Conference on Machine Learning (ICML)},
  year={2024},
  organization={PMLR}
}

@inproceedings{nakkiran-mleval22,
  author       = {P. Nakkiran and M. Belkin},
  title        = {Incentivizing Empirical Science In Machine Learning: Problems And Proposals},
  booktitle={ML Evaluation Standards Workshop at ICLR 2022},
  year={2022},
}

@article{rajwar-air23,
  author       = {K. Rajwar and
                  K. Deep and
                  S. Das},
  title        = {An exhaustive review of the metaheuristic algorithms for search and
                  optimization: taxonomy, applications, and open challenges},
  journal      = {Artif. Intell. Rev.},
  volume       = {56},
  number       = {11},
  pages        = {13187--13257},
  year         = {2023},
}

@inproceedings{crisan-facct22a,
  author       = {Anamaria Crisan and
                  Margaret Drouhard and
                  Jesse Vig and
                  Nazneen Rajani},
  title        = {Interactive Model Cards: {A} Human-Centered Approach to Model Documentation},
  booktitle    = {Proceedings of the 2022 {ACM} Conference on Fairness, Accountability, and
                  Transparency ({FAccT'22})},
  pages        = {427--439},
  publisher    = {{ACM}},
  year         = {2022}
}

@inproceedings{mitchell-fat19a,
  author       = {Margaret Mitchell and
                  Simone Wu and
                  Andrew Zaldivar and
                  Parker Barnes and
                  Lucy Vasserman and
                  Ben Hutchinson and
                  Elena Spitzer and
                  Inioluwa Deborah Raji and
                  Timnit Gebru},
  editor       = {danah boyd and
                  Jamie H. Morgenstern},
  title        = {Model Cards for Model Reporting},
  booktitle    = {Proceedings of the Conference on Fairness, Accountability, and Transparency,
                  FAT* 2019},
  pages        = {220--229},
  publisher    = {{ACM}},
  year         = {2019},
}

@software{uv,
    title = {uv: An extremely fast Python package installer and resolver},
    author = {{Astral Software Foundation}},
    year = {2024},
    url = {https://github.com/astral-sh/uv},
    organization = {Astral Software Foundation}
}

@software{ruff,
    title = {Ruff: An extremely fast Python linter and code formatter, written in Rust},
    author = {{Astral Software Foundation}},
    year = {2022},
    url = {https://github.com/astral-sh/ruff},
    organization = {Astral Software Foundation}
}

@software{black,
    title = {Black: The uncompromising Python code formatter},
    author = {Ł. Langa and contributors to Black},
    year = {2018},
    url = {https://github.com/psf/black},
}

@software{mypy,
    title = {Mypy: Static Typing for Python},
    author = {J. Lehtosalo and G. van Rossum and I. Levkivskyi and M. Sullivan},
    year = {2012},
    url = {https://github.com/python/mypy},
}

@software{pydocstyle,
    title = {pydocstyle - docstring style checker},
    author = {V. Keleshev and A. Rachum and S.Kothari},
    year = {2016},
    url = {https://github.com/PyCQA/pydocstyle},
}

@software{precommit,
    title = {pre-commit: A framework for managing and maintaining multi-language pre-commit hooks.},
    author = {A. Sottile and K. Struys and C. Kuehl},
    year = {2014},
    url = {https://pre-commit.com/},
}

@inproceedings{dewancker2016strategy,
  title={A strategy for ranking optimization methods using multiple criteria},
  author={Dewancker, Ian and McCourt, Michael and Clark, Scott and Hayes, Patrick and Johnson, Alexandra and Ke, George},
  booktitle={Workshop on Automatic Machine Learning},
  pages={11--20},
  year={2016},
  organization={PMLR}
}

@inproceedings{pfisterer2021learning,
  title={Learning multiple defaults for machine learning algorithms},
  author={Pfisterer, Florian and Van Rijn, Jan N and Probst, Philipp and M{\"u}ller, Andreas C and Bischl, Bernd},
  booktitle={Proceedings of the genetic and evolutionary computation conference companion},
  pages={241--242},
  year={2021}
}

@article{zoller2023xautoml,
  title={XAutoML: a visual analytics tool for understanding and validating automated machine learning},
  author={Z{\"o}ller, Marc-Andr{\'e} and Titov, Waldemar and Schlegel, Thomas and Huber, Marco F},
  journal={ACM Transactions on Interactive Intelligent Systems},
  volume={13},
  number={4},
  pages={1--39},
  year={2023},
  publisher={ACM New York, NY}
}

@inproceedings{anastacio2019exploitation,
  title={Exploitation of default parameter values in automated algorithm configuration},
  author={Anastacio, Marie and Luo, Chuan and Hoos, Holger},
  booktitle={Workshop Data Science meets Optimisation (DSO), IJCAI},
  volume={138},
  year={2019}
}

@ARTICLE{IOHprofiler,
  author = {Carola Doerr and Hao Wang and Furong Ye and Sander van Rijn and Thomas B{\"a}ck},
  title = {IOHprofiler: A Benchmarking and Profiling Tool for Iterative Optimization Heuristics},
  journal = {arXiv e-prints:1810.05281},
  archivePrefix = {arXiv},
  eprint = {1810.05281},
  year = {2018},
  month = {10},
  url = {https://arxiv.org/abs/1810.05281}
}

@software{copilot,
  author       = {{GitHub, Inc.}},
  title        = {GitHub Copilot: AI Pair Programmer},
  year         = {2025},
  url          = {https://docs.github.com/en/copilot},
  note         = {Software, accessed: 2025-11-21},
}

@software{cursor,
  author       = {{Cursor}},
  title        = {Cursor: AI-Assisted Code Editor},
  year         = {2025},
  url          = {https://www.cursor.com},
  note         = {Software, accessed: 2025-11-21},
}

@article{bischl-patterns25a,
    author = "Bischl, Bernd and Casalicchio, Giuseppe and Das, Taniya and Feurer, Matthias and Fischer, Sebastian and Gijsbers, Pieter and Mukherjee, Subhaditya and Müller, Andreas C. and Németh, László and Oala, Luis and Purucker, Lennart and Ravi, Sahithya and van Rijn, Jan N. and Singh, Prabhant and Vanschoren, Joaquin and van der Velde, Jos and Wever, Marcel",
    title = "OpenML: Insights from 10 years and more than a thousand papers",
    journal = "Patterns",
    volume = "6",
    year = "2025",
    doi = "http://doi.org/10.1016/j.patter.2025.101317",
    number = "7"
}

@article{benjamins-arxiv25a,
      title={carps: A Framework for Comparing N Hyperparameter Optimizers on M Benchmarks}, 
      author={Carolin Benjamins and Helena Graf and Sarah Segel and Difan Deng and Tim Ruhkopf and Leona Hennig and Soham Basu and Neeratyoy Mallik and Edward Bergman and Deyao Chen and François Clément and Matthias Feurer and Katharina Eggensperger and Frank Hutter and Carola Doerr and Marius Lindauer},
      year={2025},
      eprint={2506.06143},
      archivePrefix={arXiv},
      primaryClass={cs.LG},
      url={https://arxiv.org/abs/2506.06143}, 
}

@inproceedings{tschalzev-fmldpr25a,
title = {Unreflected Use of Tabular Data Repositories Can Undermine Research Quality},
author = {Andrej Tschalzev and Lennart Purucker and Stefan Lüdtke and Frank Hutter and Christian Bartelt and Heiner Stuckenschmidt},
year = {2025},
booktitle = {The Future of Machine Learning Data Practices and Repositories at ICLR}
}

@inproceedings{schneider-automl25a,
    author = {L. Schneider and B. Bischl and M. Feurer},
    title = {Overtuning in Hyperparameter Optimization},
    booktitle = {To appear in the proceedings of the 4th International Conference on Automated Machine Learning},
    year = {2025},
}

@article{gijsbers-jmlr24a,
  author  = {Pieter Gijsbers and Marcos L. P. Bueno and Stefan Coors and Erin LeDell and S{{\'e}}bastien Poirier and Janek Thomas and Bernd Bischl and Joaquin Vanschoren},
  title   = {AMLB: an AutoML Benchmark},
  journal = {Journal of Machine Learning Research},
  year    = {2024},
  volume  = {25},
  number  = {101},
  pages   = {1--65},
  url     = {http://jmlr.org/papers/v25/22-0493.html}
}

@inproceedings{margraf2025synthacticbench,
  title={SynthACticBench: A Capability-Based Synthetic Benchmark for Algorithm Configuration},
  author={Margraf, Valentin and Lappe, Anna and Wever, Marcel and Benjamins, Carolin and H{\"u}llermeier, Eyke and Lindauer, Marius},
  booktitle={Proceedings of the Genetic and Evolutionary Computation Conference},
  pages={39--47},
  year={2025}
}

@inproceedings{bossek2019evolving,
  title={Evolving diverse TSP instances by means of novel and creative mutation operators},
  author={Bossek, Jakob and Kerschke, Pascal and Neumann, Aneta and Wagner, Markus and Neumann, Frank and Trautmann, Heike},
  booktitle={Proceedings of the 15th ACM/SIGEVO conference on foundations of genetic algorithms},
  pages={58--71},
  year={2019}
}

@online{Liefooghe2023mocobench,
  author    = {Arnaud Liefooghe},
  title     = {MoCObench},
  year      = {2023},
  month     = jul,
  day       = {21},
  url       = {https://gitlab.com/aliefooghe/mocobench},
  note      = {Source code repository},
  organization = {GitLab},
}

@article{lopez2021reproducibility,
  title={Reproducibility in evolutionary computation},
  author={L{\'o}pez-Ib{\'a}{\~n}ez, Manuel and Branke, Juergen and Paquete, Lu{\'\i}s},
  journal={ACM Transactions on Evolutionary Learning and Optimization},
  volume={1},
  number={4},
  pages={1--21},
  year={2021},
  publisher={ACM New York, NY}
}

@article{hansen-report09,
    author = {N. Hansen and A. Auger and S. Finck and R. Ros},
    title = {Real-Parameter Black-Box Optimization Benchmarking 2009: Experimental Setup},
  journal={Technical Report RR-6828},
  year={2009},
  publisher={INRIA}
}

@article{lessing-69,
    author = {G.E. Lessing},
    title = {The Hamburg Dramaturgy},
    year={1769},
}

@article{liu-acm23,
  author       = {P. Liu and
                  W. Yuan and
                  J. Fu and
                  Z. Jiang and
                  H. Hayashi and
                  G. Neubig},
  title        = {Pre-train, Prompt, and Predict: {A} Systematic Survey of Prompting
                  Methods in Natural Language Processing},
  journal      = {{ACM} Comput. Surv.},
  volume       = {55},
  number       = {9},
  pages        = {195:1--195:35},
  year         = {2023},
  doi          = {10.1145/3560815},
}

@inproceedings{fernando-iclr24,
  author       = {C. Fernando and
                  D. Banarse and
                  H. Michalewski and
                  S. Osindero and
                  T. Rockt{\"{a}}schel},
  title        = {Promptbreeder: Self-Referential Self-Improvement via Prompt Evolution},
  booktitle    = {Forty-first International Conference on Machine Learning, {ICML}},
  publisher    = {OpenReview.net},
  year         = {2024},
  url          = {https://openreview.net/forum?id=9ZxnPZGmPU},
}

@inproceedings{wang-iclr23,
  author       = {Z. Wang and
                  R. Panda and
                  L. Karlinsky and
                  R. Feris and
                  H. Sun and
                  Y. Kim},
  title        = {Multitask Prompt Tuning Enables Parameter-Efficient Transfer Learning},
  booktitle    = {The Eleventh International Conference on Learning Representations,
                  {ICLR} 2023},
  publisher    = {OpenReview.net},
  year         = {2023},
  url          = {https://openreview.net/forum?id=Nk2pDtuhTq},
}

@inproceedings{petelin-gecco25,
  author       = {G. Petelin and
                  G. Cenikj},
  title        = {The Pitfalls of Benchmarking in Algorithm Selection: What We Are Getting
                  Wrong},
  booktitle    = {Proceedings of the Genetic and Evolutionary Computation Conference,
                  {GECCO}},
  pages        = {1181--1189},
  publisher    = {{ACM}},
  year         = {2025},
  doi          = {10.1145/3712256.3726336},
}

@inproceedings{jordan-icml20,
  author       = {E. Jordan and
                  Y. Chandak and 
                  D. Cohen and 
                  M. Zhang and
                  P. Thomas },
  title        = {Evaluating the Performance of Reinforcement Learning Algorithms},
  crossref = {icml20}
}

@article{kerr1998harking,
  title={HARKing: Hypothesizing after the results are known},
  author={Kerr, Norbert L},
  journal={Personality and social psychology review},
  volume={2},
  number={3},
  pages={196--217},
  year={1998},
  publisher={Sage Publications Sage CA: Los Angeles, CA}
}

@inproceedings{schneider-ppsn22b,
  title        = {HPO $\times$ ELA: Investigating Hyperparameter Optimization Landscapes by Means of Exploratory Landscape Analysis},
  author       = "Schneider, Lennart and Sch{\"a}permeier, Lennart and Prager, Raphael Patrick and Bischl, Bernd and Trautmann, Heike and Kerschke, Pascal",
  year         = {2022},
  pages        = {575--589},
  crossref     = {ppsn22},
}

@article{liu-air25,
  author       = {Y. Liu and
                  J. Cao and
                  C. Liu and
                  K. Ding and
                  L. Jin},
  title        = {Datasets for large language models: a comprehensive survey},
  journal      = {Artif. Intell. Rev.},
  volume       = {58},
  number       = {12},
  pages        = {403},
  year         = {2025},
  doi          = {10.1007/S10462-025-11403-7},
}

@article{kaplan-corr20,
  author       = {J. Kaplan and
                  S. McCandlish and
                  T. Henighan and
                  T. Brown and
                  B. Chess and
                  R. Child and
                  S. Gray and
                  A. Radford and
                  J. Wu and
                  D. Amodei},
  title        = {Scaling Laws for Neural Language Models},
  journal      = {CoRR},
  volume       = {abs/2001.08361},
  year         = {2020},
}

@inproceedings{magar-acl22,
  author       = {I. Magar and
                  R. Schwartz},
  title        = {Data Contamination: From Memorization to Exploitation},
  booktitle    = {Proceedings of the 60th Annual Meeting of the Association for Computational
                  Linguistics (Volume 2: Short Papers), {ACL}},
  pages        = {157--165},
  publisher    = {Association for Computational Linguistics},
  year         = {2022},
  doi          = {10.18653/V1/2022.ACL-SHORT.18},
}

@inproceedings{balloccu-eacl24,
  author       = {S. Balloccu and
                  P. Schmidtov{\'{a}} and
                  M. Lango and
                  O. Dusek},
  editor       = {Yvette Graham and
                  Matthew Purver},
  title        = {Leak, Cheat, Repeat: Data Contamination and Evaluation Malpractices
                  in Closed-Source LLMs},
  booktitle    = {Proceedings of the 18th Conference of the European Chapter of the
                  Association for Computational Linguistics, {EACL}},
  pages        = {67--93},
  publisher    = {Association for Computational Linguistics},
  year         = {2024},
}

@inproceedings{sainz-conda24,
  author       = {O. Sainz and
                  I. Garc{\'{\i}}a{-}Ferrero and
                  A. Jacovi and
                  J. Campos and
                  Y. Elazar and
                  E. Agirre and
                  Y. Goldberg and
                  W. Chen and
                  J. Chim and
                  L. Choshen and
                  L. D'Amico{-}Wong and
                  M. Dell and
                  R. Fan and
                  S. Golchin and
                  Y. Li and
                  P. Liu and
                  B. Pahwa and
                  A. Prabhu and
                  S. Sharma and
                  E. Silcock and
                  K. Solonko and
                  D. Stap and
                  M. Surdeanu and
                  Y. Tseng and
                  V. Udandarao and
                  Z. Wang and
                  R. Xu and
                  J. Yang},
  title        = {Data Contamination Report from the 2024 {CONDA} Shared Task},
  booktitle    = {The First Data Contamination Workshop, {CONDA}},
  pages        = {41--56},
  publisher    = {Association for Computational Linguistics},
  year         = {2024},
}

@article{thakkar-corr25,
  author       = {N. Thakkar and
                  M. Y{\"{u}}ksekg{\"{o}}n{\"{u}}l and
                  J. Silberg and
                  A. Garg and
                  N. Peng and
                  F. Sha and
                  R. Yu and
                  C. Vondrick and
                  J. Zou},
  title        = {Can {LLM} feedback enhance review quality? {A} randomized study of
                  20K reviews at {ICLR} 2025},
  journal      = {CoRR},
  volume       = {abs/2504.09737},
  year         = {2025},
  doi          = {10.48550/ARXIV.2504.09737},
}

@inproceedings{kohli-dmlr24,
  author       = {R. Kohli and
                  M. Feurer and
                  K. Eggensperger and
                  B. Bischl and
                  F. Hutter},
  title        = {Towards Quantifying the Effect of Datasets for Benchmarking:
A Look at Tabular Machine Learning},
  booktitle    = {Data-centric Machine Learning Workshop at {ICLR}},
  year         = {2024},
}

@InProceedings{pmlr-v188-pfisterer22a,
  title = 	 {YAHPO Gym - An Efficient Multi-Objective Multi-Fidelity Benchmark for Hyperparameter Optimization},
  author =       {Pfisterer, Florian and Schneider, Lennart and Moosbauer, Julia and Binder, Martin and Bischl, Bernd},
  booktitle = 	 {Proceedings of the First International Conference on Automated Machine Learning},
  pages = 	 {3/1--39},
  year = 	 {2022},
  editor = 	 {Guyon, Isabelle and Lindauer, Marius and van der Schaar, Mihaela and Hutter, Frank and Garnett, Roman},
  volume = 	 {188},
  series = 	 {Proceedings of Machine Learning Research},
  month = 	 {25--27 Jul},
  publisher =    {PMLR},
  url = 	 {https://proceedings.mlr.press/v188/pfisterer22a.html}
}

@inproceedings{vermetten2024largescale,
author = {Vermetten, Diederick and Doerr, Carola and Wang, Hao and Kononova, Anna V. and B\"{a}ck, Thomas},
title = {Large-Scale Benchmarking of Metaphor-Based Optimization Heuristics},
year = {2024},
isbn = {9798400704949},
publisher = {Association for Computing Machinery},
address = {New York, NY, USA},
url = {https://doi.org/10.1145/3638529.3654122},
doi = {10.1145/3638529.3654122},
booktitle = {Proceedings of the Genetic and Evolutionary Computation Conference},
pages = {41–49},
numpages = {9},
location = {Melbourne, VIC, Australia},
series = {GECCO '24}
}

@proceedings{aaai15,
  title        = {Proceedings of the Twenty-ninth {AAAI} Conference on Artificial Intelligence ({AAAI}'15)},
  year         = 2015,
  booktitle    = {Proceedings of the Twenty-ninth {AAAI} Conference on Artificial Intelligence ({AAAI}'15)},
  publisher    = {{AAAI} Press},
  editor       = {B. Bonet and S. Koenig},
}

@proceedings{aistats17,
  title        = {Proceedings of the Seventeenth International Conference on Artificial Intelligence and Statistics ({AISTATS}'17)},
  year         = 2017,
  booktitle    = {Proceedings of the Seventeenth International Conference on Artificial Intelligence and Statistics ({AISTATS}'17)},
  publisher    = {Proceedings of Machine Learning Research},
  volume       = 54,
  editor       = {A. Singh and J. Zhu},
}

@proceedings{iclr22,
  title        = {Proceedings of the International Conference on Learning Representations ({ICLR}'22)},
  year         = 2022,
  booktitle    = {Proceedings of the International Conference on Learning Representations ({ICLR}'22)},
  note         = {Published online: \url{iclr.cc}},
}

@proceedings{icml19,
  title        = {Proceedings of the 36th International Conference on Machine Learning ({ICML}'19)},
  year         = 2019,
  booktitle    = {Proceedings of the 36th International Conference on Machine Learning ({ICML}'19)},
  publisher    = {Proceedings of Machine Learning Research},
  volume       = 97,
  editor       = {K. Chaudhuri and R. Salakhutdinov},
}

@proceedings{icml20,
  title        = {Proceedings of the 37th International Conference on Machine Learning ({ICML}'20)},
  year         = 2020,
  booktitle    = {Proceedings of the 37th International Conference on Machine Learning ({ICML}'20)},
  publisher    = {Proceedings of Machine Learning Research},
  volume       = 98,
  editor       = {H. Daume III and A. Singh},
}

@proceedings{ijcai21,
  title        = {Proceedings of the 30th International Joint Conference on Artificial Intelligence ({IJCAI}'21)},
  year         = 2021,
  booktitle    = {Proceedings of the 30th International Joint Conference on Artificial Intelligence ({IJCAI}'21)},
  editor       = {Z. Zhou},
}

@proceedings{lion18,
  title        = {Proceedings of the International Conference on Learning and Intelligent Optimization ({LION})},
  year         = 2018,
  booktitle    = {Proceedings of the International Conference on Learning and Intelligent Optimization ({LION})},
  publisher    = springer,
  series       = lncs,
  editor       = {R. Battiti and M. Brunato and I. Kotsireas and P. Pardalos},
}

@proceedings{mlsys21,
  title        = {Proceedings of Machine Learning and Systems 3},
  year         = 2021,
  booktitle    = {Proceedings of Machine Learning and Systems 3},
  volume       = 3,
  editor       = {A. Smola and A. Dimakis and I. Stoica},
}

@proceedings{neuripsdbt21,
  title        = {Proceedings of the Neural Information Processing Systems Track on Datasets and Benchmarks},
  year         = 2021,
  booktitle    = {Proceedings of the Neural Information Processing Systems Track on Datasets and Benchmarks},
  publisher    = curran,
  editor       = {J. Vanschoren and S. Yeung},
}

@proceedings{ppsn22,
  title        = {Proceedings of the Sixteenth International Conference on Parallel Problem Solving from Nature ({PPSN}'22)},
  year         = 2022,
  booktitle    = {Proceedings of the Sixteenth International Conference on Parallel Problem Solving from Nature ({PPSN}'22)},
  publisher    = springer,
  series       = lncs,
  editor       = {G. Rudolph and A. V. Kononova and H. Aguirre and P. Kerschke and G. Ochoa and T. Tusar},
}

@proceedings{realml22,
  title        = {{ICML} Workshop on Adaptive Experimental Design and Active Learning in the Real World (ReALML Workshop 2022)},
  year         = 2022,
  booktitle    = {{ICML} Adaptive Experimental Design and Active Learning in the Real World (ReALML Workshop 2022)},
  editor       = {M. Mutny and I. Bogunovic and W. Neiswanger and S. Ermon and Y. Yue and A. Krause},
}

@proceedings{sat21,
  title        = {Proceedings of the Twenty-fourth International Conference on Theory and Applications of Satisfiability Testing ({SAT}'21)},
  year         = 2021,
  booktitle    = {Proceedings of the Twenty-fourth International Conference on Theory and Applications of Satisfiability Testing ({SAT}'21)},
  publisher    = springer,
  series       = lncs,
  editor       = {C. Li and F. Many{\`{a}}},
}

@STRING{acm     = "ACM Press" }

@STRING{aaai    = "Proceedings of the National Conference on Artificial
                  Intelligence (AAAI)" }

@STRING{ai      = "Artificial Intelligence" }

@STRING{aij     = "Artificial Intelligence" }

@STRING{amai    = "Annals of Mathematics and Artificial Intelligence" }

@STRING{curran  = "Curran Associates" }

@STRING{ieee = "IEEE" }

@STRING{ijcai   = "Proceedings of the International Joint Conference on
                  Artificial Intelligence" }

@STRING{inria   = "INRIA" }

@STRING{is      = "Informatik-Spektrum" }

@STRING{jair    = "Journal of Artificial Intelligence Research" }

@STRING{jmlr    = "Journal of Machine Learning Research" }

@STRING{kaufmann = "Morgan Kaufmann Publishers" }

@STRING{kluwer  = "Kluwer Academic Publishers" }

@STRING{lncs    = "Lecture Notes in Computer Science" }

@STRING{nature = "Nature"}

@STRING{plos    = "Public Library of Science San Francisco, CA USA"}

@STRING{pmlr    = "Proceedings of Machine Learning Research"}

@STRING{springer = "Springer" }

@STRING{wiley   = "John Wiley \& sons" }
\end{document}